\pdfoutput=1
\documentclass[11pt]{article}

\usepackage[final]{acl}

\usepackage{times}
\usepackage{latexsym}

\usepackage[T1]{fontenc}
\usepackage[utf8]{inputenc}

\usepackage{microtype}

\usepackage{inconsolata}

\usepackage{booktabs}
\usepackage{amsfonts}
\usepackage{nicefrac}
\usepackage{microtype}
\usepackage[pdftex]{graphicx}
\usepackage{float}
\usepackage{enumitem}
\usepackage{subcaption}
\usepackage[export]{adjustbox}
\usepackage{amsmath}
\usepackage{tikz}
\usepackage[most]{tcolorbox}
\usepackage{caption}
\usepackage{array}
\usepackage{soul}
\usepackage[super]{nth}
\usepackage{float}
\usepackage{wrapfig}
\usepackage{mdframed}
\usepackage{multirow}

\usepackage{multicol}
\usepackage{algorithm}
\usepackage{listings}
\usepackage{inconsolata}
\usepackage{times}
\usepackage{latexsym}
\usepackage{bm}
\usepackage{amsthm,amsmath,amssymb}
\usepackage{mathrsfs}
\usepackage{epsfig}
\usepackage{booktabs}
\usepackage{verbatim}
\usepackage{enumitem,kantlipsum}
\usepackage{lipsum}
\usepackage{tipa}
\usepackage{algorithmicx}
\usepackage{algpseudocode}
\usepackage{multirow,makecell}
\usepackage{arydshln}
\usepackage{colortbl}
\usepackage{caption}
\usepackage{hyperref}
\usepackage{color} 
\usepackage{dsfont}
\usepackage{tikz}
\usepackage{wrapfig}
\usepackage{multirow} 
\usepackage{xspace}
\usepackage{ifthen}
\usepackage{caption}
\usepackage{subcaption}
\usepackage{pdflscape}
\usepackage{pifont}
\newcommand{\xmark}{\ding{55}}
\newcommand{\cmark}{\ding{51}}
\newtcolorbox{AIbox}[2][]{aibox,title=#2,#1}
\tcbset{
  aibox/.style={
    width=\textwidth,
    top=10pt,
    colframe=black,
    colbacktitle=black,
    enhanced,
    center,
    attach boxed title to top left={yshift=-0.1in,xshift=0.15in},
    boxed title style={boxrule=0pt,colframe=white,},
  }
}
\definecolor{forestgreen}{rgb}{0.13, 0.55, 0.13}

\definecolor{gblue}{HTML}{4285F4}
\definecolor{gred}{HTML}{DB4437}
\definecolor{ggreen}{HTML}{0F9D58}

\definecolor{mygray}{gray}{.92}

\newcommand{\ours}{\textsc{TextBind}\xspace}
\newcommand{\ourmodel}{\textsc{MIM}\xspace}
\newcommand{\ourbench}{\textsc{TextBindEval}\xspace}

\newcommand{\rev}[1]{\textcolor{black}{#1}}

\title{\ours: \\ Multi-turn Interleaved Multimodal Instruction-following in the Wild}
\author{
Huayang Li$^{\heartsuit\spadesuit}$\thanks{Equal Contribution. Work done during HL and SL's internships at Tencent AI Lab. Correspondence to DC (\texttt{\url{thisisjcykcd@gmail.com}}).}\ \ \ \ \ Siheng Li$^{\heartsuit\clubsuit}$\footnotemark[1]\ \ \ \ \ Deng Cai$^{\heartsuit}$\footnotemark[1] \ \ \ \ \ Longyue Wang$^{\heartsuit}$ \AND Lemao Liu$^{\heartsuit}$\ \ \ \ \ Taro Watanabe$^\spadesuit$\ \ \ \ \ Yujiu Yang$^\clubsuit$\ \ \ \ \ Shuming Shi$^{\heartsuit}$\\
  \\
   $^\heartsuit$Tencent AI Lab\ \ \ \ \ $^\spadesuit$Nara Institute of Science and Technology\ \ \ \ \
   $^\clubsuit$Tsinghua University\\
  \\
}

\begin{document}
\maketitle
\begin{abstract}
Large language models with instruction-following abilities have revolutionized the field of artificial intelligence. These models show exceptional generalizability to tackle various real-world tasks through their natural language interfaces. However, their performance heavily relies on high-quality exemplar data, which is often difficult to obtain. This challenge is further exacerbated when it comes to multimodal instruction following. We introduce \ours, an almost annotation-free framework for empowering LLMs with multi-turn interleaved multimodal instruction-following capabilities. Our approach requires only image-caption pairs and generates multi-turn multimodal instruction-response conversations from a language model. To accommodate interleaved image-text inputs and outputs, we devise \ourmodel, a language model-centric architecture that seamlessly integrates image encoder and decoder models. Extensive quantitative and qualitative experiments demonstrate that \ourmodel trained on \ours achieves remarkable generation capability in multimodal conversations compared to recent baselines.
%We release our dataset, model, and demo to foster future research in the area of multimodal instruction following.
\end{abstract}

\section{Introduction}

Artificial intelligence (AI) has experienced a significant paradigm shift with the rise of large language models (LLMs). These models are capable of processing a wide range of natural language processing (NLP) applications through natural language interactions with users \citep{chatgpt,OpenAI2023GPT4TR}.
%Despite their remarkable performance, these models cannot process and generate visual content.
%Building an interactive virtual assistant that can effectively process both vision and language information is of great importance and utility.

Recently, a number of efforts have been made to augment LLMs with visual perception and understanding abilities. Prior work uses template-based instruction-following datasets for training \citep{xu-etal-2023-multiinstruct,dai2023instructblip,li2023m}. These datasets comprise a variety of classic computer vision (CV) tasks, e.g., object detection, with each task being converted into an instructional format using a handful of human-written natural language instructions. However, classic CV tasks often represent manageable and focused abstractions or simplifications of real-world tasks \citep{marr2010vision}, they generally fall short in representing the true variety and complexity of real-world tasks and capturing the lexical diversity of human language. For example, most of them are single-turn inquiries about a single input image, albeit a small fraction supports multi-turn textual interactions or multiple image inputs. Consequently, the instruction-following capabilities of models trained on these datasets remain limited in open-world scenarios \citep{xu2023lvlm}. This is reminiscent of the early development of instruction tuning in NLP, where public NLP tasks were initially adopted but eventually superseded by high-quality, diverse open-world instruction data \citep{ouyang2022training}. Nevertheless, collecting such data for multimodal models can be extremely costly. 

In this paper, we address the above challenge by introducing \ours, an almost annotation-free framework for augmenting LLMs with multi-turn interleaved multimodal instruction-following capabilities. The main idea is to represent images through their textual descriptions, e.g., captions, and utilize an LLM to generate multi-turn instructions and responses. To ensure the coherence and meaningfulness of the constructed multi-turn conversations, we propose a series of strategies such as topic-aware image sampling and human-in-the-loop refinement of in-context demonstrations. \ours can harvest large-scale datasets given the abundance of public image-caption pairs. \ours provides examples of processing and generating arbitrarily interleaved image-text content. To accommodate interleaved image-text inputs and outputs, we devise \ourmodel, a multimodal model that emphasizes the reasoning abilities of LLMs and seamlessly integrates image encoder and decoder models. 
% Most interestingly, the core innovation of our model is its capability to interact with users naturally. For instance, rather than requiring users to supply the model with explicit descriptions of the desired image, our model can spontaneously generate images in proper conversation contexts.
% Our qualitative experiments show that \ourmodel trained on \ours can perform a wide range of tasks, including composing engaging stories inspired by a set of images, identifying commonalities and differences across multiple images, explaining concepts with vivid images, generating long coherent stories with illustrations, etc. Examples can be found in Appx. \ref{app:demonstration}. 

For a comprehensive evaluation of open-world multimodal instruction following, we also construct \ourbench, an evaluation set of \ours validated and refined by human annotators. To verify the effectiveness of \ours, we compare \ourmodel models trained on \ours and various existing datasets such as MiniGPT-4 \citep{zhu2023minigpt} and LLaVA \citep{liu2023llava}. The results demonstrate that \ours substantially outperforms these datasets across multiple evaluation metrics. Additional evaluations on conventional vision-language, including MME~\citep{Fu2023MMEAC}, MMBench~\citep{liu2023mmbench}, and MM-Vet~\citep{yu2023mm}, further confirm that \ours provides complementary benefits to existing multimodal datasets. For image generation quality, we compare the proposed \ourmodel with GILL \cite{koh2023generating} and SDXL \cite{podell2023sdxl}, showcasing improved performance (e.g., 0.673 vs. 0.608 CLIP scores). The advantage is particularly pronounced in multi-turn scenarios. Moreover, a holistic human evaluation further substantiates the superior performance of \ourmodel trained on \ours in contrast to baseline models in the context of open-world multi-turn interleaved multimodal instruction-following. Most interestingly, rather than requiring users to provide explicit commands and direct descriptions of the desired image, our model can spontaneously generate images in proper conversation contexts. Our qualitative experiments show that \ourmodel trained on \ours can perform a wide range of tasks, including composing engaging stories inspired by a set of images, identifying commonalities and differences across multiple images, explaining concepts with vivid images, generating long coherent stories with illustrations, etc. Examples can be found in Appx. \ref{app:demonstration}. We hope \ours serves as an initial step towards building AGI that can interact with humans flexibly in different modalities and broad real-world scenarios.

\section{Related Work}
\vspace{-2mm}
\paragraph{Multimodal Datasets}
Existing multimodal datasets can be broadly classified into two categories: (1) Conventional datasets for specific vision-language tasks such as image captioning \citep{coco_caption,nocaps,flickr30k} and visually-grounded question answering \citep{hudson2019gqa,ok-vqa,textvqa,scienceqa,youcook2,vqav2,vizwiz}. (2) Recent datasets for general instruction following. For instance, MultiInstruct \citep{xu-etal-2023-multiinstruct}, InstructBLIP \citep{dai2023instructblip}, and M3IT \citep{li2023m} convert existing vision-language datasets into a unified instructional format with handcrafted templates. This approach is reminiscent of the early explorations on instruction tuning in NLP \citep{wei2021finetuned,sanh2021multitask}, where existing NLP tasks were phrased as instructions. However, it has been reported that such instruction-tuned multimodal models still generalize poorly to open-world scenarios \citep{xu2023lvlm}. This finding also aligns with the observations in NLP \citep{ouyang2022training}, where template-based instruction tuning is less effective than instruction tuning data collected from real-world scenarios due to its restricted diversity. There are also some attempts to convert the output of existing vision-language models into natural language answers for constructing instruction-tuning data \citep{liu2023llava,zhu2023minigpt,chen2023shikra}.

Compared to existing instruction-tuning data, the examples in \ours (1) generally exhibit greater task and lexicon diversity; (2) typically involve multiple images scattered throughout a multi-urn conversation; (3) support multimodal output (image generation).
\vspace{-2mm}
\paragraph{Multimodal Models}
To augment existing LLMs with visual abilities, one straightforward approach is to employ off-the-shelf vision models as external tools. That is, the LLM calls expert vision models through their language interfaces for completing specific visual tasks when needed \citep{visual_chat_gpt,hugginggpt,cola,zou2022generalized,yang2023mm,suris2023vipergpt}.However, these approaches may suffer from cross-modal information loss and lack of generality.

Recently, end-to-end multimodal language models have garnered significant interest. Flamingo \citep{alayrac2022flamingo} and OpenFlamingo \citep{alayrac2022flamingo} are among the pioneering work to extend LLMs to vision-language pretraining. Different from training from scratch, subsequent research efforts have focused on integrating pretrained vision and language models. BLIP-2~\citep{li2023blip} proposes Qformer to align the feature spaces of vision models and language models. To date, various network architectures and training strategies have been proposed \citep{zhu2023minigpt,liu2023llava,ye2023mplug,li2023otter,llama_adapater,du2022glm,chen2023shikra,dai2023instructblip,liu2023mitigating}.
%These multimodal language models typically undergo a two-stage training process. In the first stage, image-caption data are used to align the hidden space of vision and language models. In the second stage, the joint model is trained on vision-language tasks phrased as instructions.
However, these models are limited to the use of visual content as input. Our work is inspired by recent work on LLM-empowered image retrieval or generation \citep{koh2023grounding,koh2023generating} \rev{and the pioneer work of \citep{sun-etal-2022-multimodal} for chitchat in the context of single photo sharing.} Contrary to prior work, we aim to present the first instruction-following model capable of processing and generating arbitrarily interleaved image-text inputs and outputs.

Numerous contemporary studies also exist in this field \citep{2023internlm, yao2023deepspeed-visualchat, dong2023dreamllm, zheng2023minigpt5, ge2023making}. A unique characteristic of our work lies in our emphasis on the aspect of data creation, whereas other studies primarily concentrate on architecture design or training algorithms. \citet{pan2023kosmos} and \citet{chen2023llava} focus on image editing, while \citet{wu2023nextgpt} and \citet{moon2023anymal} place emphasis on incorporating additional modalities, such as audio and video.
\paragraph{Evaluation}
Conventional vision datasets designed for specific tasks and scenarios may suffer from data contamination issues for evaluating LLMs. Recently, efforts have been made to provide systematic evaluations with a broader coverage of diverse visual abilities. MME \citep{Fu2023MMEAC} is an evaluation dataset containing visually-grounded Yes/No questions.
OwlEval \citep{ye2023mplug} is a benchmark comprising 82 questions based on 50 images and relies on human feedback evaluation. The test size is limited, and the results may suffer from subjective bias. In response to these challenges, MMbench \citep{liu2023mmbench} and MM-Vet \citep{yu2023mm} are two recent benchmarks aiming to offer more comprehensive evaluations by incorporating the use of ChatGPT/GPT4 for answer verification. LVLM Arena \citep{xu2023lvlm}, an online evaluation framework that ranks different models using human judgment, is also introduced. However, the above benchmarks primarily focus on question answering based on a single image at the beginning of a conversation.

\section{\ours \label{sec:method}}
In this work, we seek to enhance the multi-turn instruction-following capabilities of a language model in the context of arbitrarily interleaved images and text. Constructing such datasets poses significant challenges: 1) it demands inventive thinking for devising high-quality visually-grounded instructions and their responses; 2) it requires specialized expertise to craft appropriate images. To tackle these issues, we introduce \ours, a method that predominantly resorts to existing \textit{text-only language models} to produce the desired data.

\subsection{Definition of Data\label{sec:def}}
The goal of \ours is to construct a collection of multi-turn conversation such as $[\boldsymbol{x}^{1}_{u}, \boldsymbol{x}^{1}_{a},\ldots,\boldsymbol{x}^{T}_{u}, \boldsymbol{x}^{T}_{a}]$, where $T$ is the number of turns, $\boldsymbol{x}^{i}_{u}$ denotes the $i$-th instruction from the user, and $\boldsymbol{x}^{i}_{a}$ represents the $i$-th response from the assistant. The conversation is also accompanied by an image set $\{\boldsymbol{m}_{1}, \ldots, \boldsymbol{m}_{n} \}$, where $n$ is the number of unique images in this conversation. Each instruction $\boldsymbol{x}^{i}_{u}$ or response $\boldsymbol{x}^{i}_{a}$ is a sequence of tokens in $\mathcal{V}_{\text{lang}} \cup \mathcal{V}_{\text{img}}$, where $\mathcal{V}_{\text{lang}}$ is the ordinary vocabulary of a language model and $\mathcal{V}_{\text{img}}$ contains $n$ distinct pointers to the images $\boldsymbol{m}_{1}, \ldots, \boldsymbol{m}_{n}$ respectively. It is worth noting that every image can appear at any point within the conversation.
\subsection{Automatic Data Generation}

%Our method capitalizes on the powerful language understanding and generation abilities of LLMs as effective annotators. However, LLM cannot "see". To circumvent the problem, we use image captions as textualized proxies of images and perform data construction entirely in language space. This flexibility allows for the creation of arbitrarily interleaved text-and-image content.
%Another concern is that a random image set may result in incoherent and unrealistic conversations.
% The core idea of \ours involves generating the information of conversations entirely in language space and subsequently mapping the textualized visual information back to the corresponding images. Unlike previous studies that provide a single image at the beginning of a conversation \citep{liu2023visual, zhu2023minigpt}, our desired conversation structure involves multiple images interspersed with text. Randomly sampling images for a conversation may lead to incoherent and unrealistic conversations. 

\begin{figure*}[t]
    \begin{center}
        \includegraphics[width=0.90\linewidth]{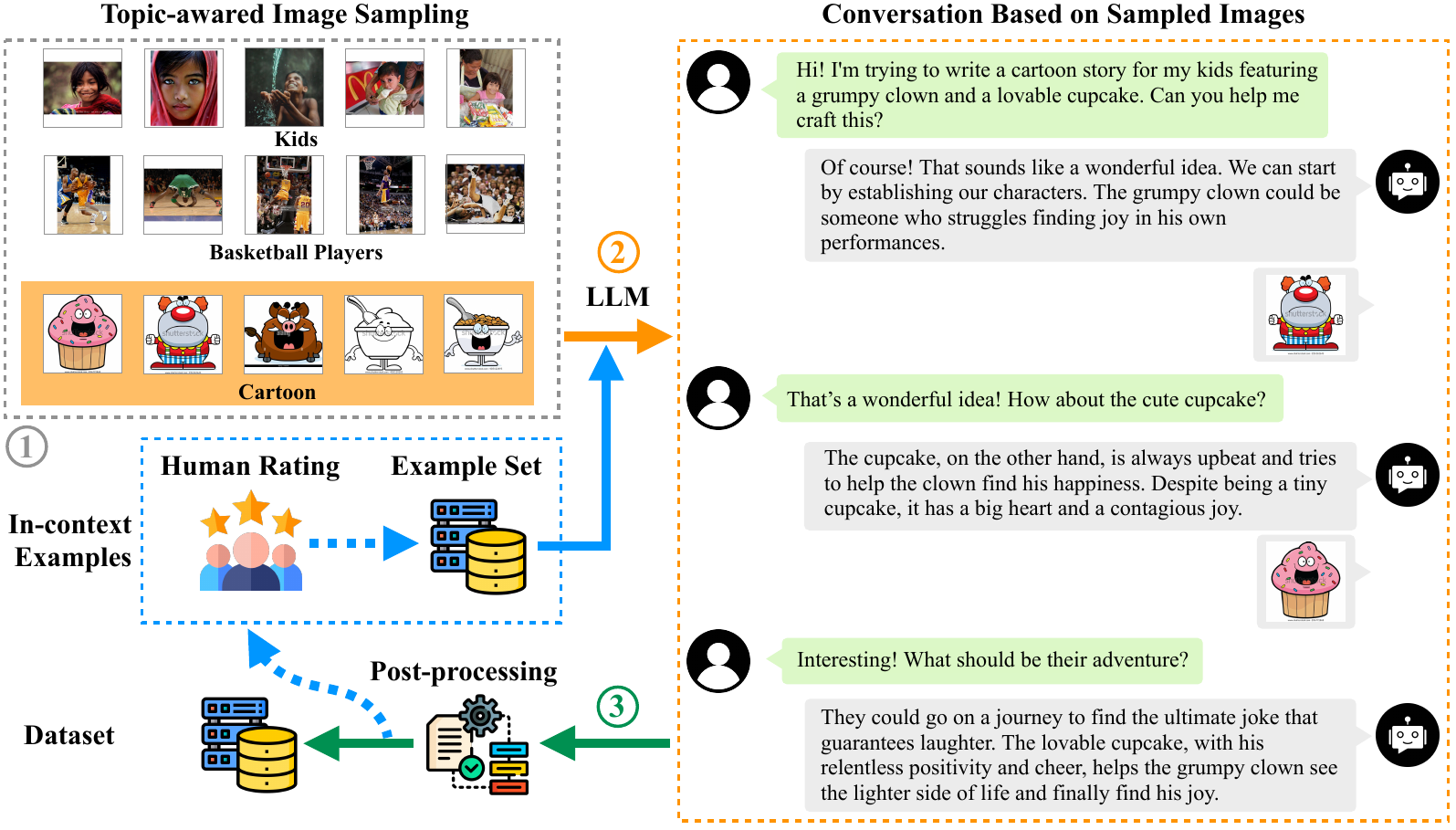}
    \end{center}
        \caption{Illustration of the \ours method. In the top-left corner, we display five representative images from each of the three example clusters obtained via unsupervised clustering. On the right-hand side, a conversation is showcased and constructed using two randomly sampled images from the cartoon cluster.%In the bottom-left corner, we outline the additional \ours pipeline, which includes human-in-the-loop refinement and post-processing stages.
\label{fig:illustration}}
\end{figure*}

\ours consists of a three-step pipeline: 1) topic-aware image sampling for ensuring the coherence of each conversation and the diversity across conversations; 2) LLM-empowered multi-turn instruction-response generation to create natural and practical conversations; 3) post-processing and filtering to eliminate low-quality data. An overview of the pipeline is shown in Fig. \ref{fig:illustration}.
\vspace{-2mm}
\paragraph{Topic-Aware Image Sampling}
The initial step of \ours entails assembling groups of images that will serve as the foundation for generating multi-turn conversations. In order to facilitate coherent, meaningful, and practical conversations, the images within each group should exhibit meaningful interconnections. Furthermore, to guarantee a comprehensive representation of real-world scenarios, the topics of images across different conversations should demonstrate a wide range of diversity.

Following the above inspirations, we employ unsupervised clustering algorithms to group the images in our dataset into clusters and execute a two-step image sampling process for each conversation. Concretely, we use the image encoder of the CLIP model \citep{pmlr-v139-radford21a} to obtain vector representations of images. Then, we execute the $k$-means algorithm to classify all images into $K$ clusters (topics). Examples of such clusters are given in Fig. \ref{fig:illustration}. For each conversation, we randomly sample a cluster from the available $K$ clusters, then sample \rev{$n \in \{2,3,4\}$} images from the chosen cluster. \rev{We want to higlight that the clustered images are semantically relevant, rather than visually similar.}
\vspace{-2mm}
\paragraph{Generation of Multi-turn Conversations}
%\footnote{Although OpenAI claims that GPT4 supports visual input, this feature is yet to be made public at the time of our research.}
After selecting a list of images, we proceed to leverage a text-only LLM, such as GPT-4, to simulate a conversation between a user and an assistant based on the chosen images. The core idea is to let LLMs receive and process the textual descriptions of the images as if they see the actual images. Given the abundance of publicly available image-caption pairs, we propose representing an image with an XML-like string \texttt{<imgX> DESCRIPTION </imgX>}, where \texttt{DESCRIPTION} serves as a placeholder for the image caption, \texttt{<imgX>} and \texttt{</imgX>} mark the caption boundaries, and \texttt{X} denotes the image index in the input image list. After generating the conversation, we replace the XML-like strings in the conversation with the original images. Importantly, to ensure that a caption faithfully describes its corresponding image, we employ the CLIP model \citep{pmlr-v139-radford21a} to filter out image-caption pairs with matching scores below a high threshold.

The detailed prompt can be found in Appx. \ref{app:gpt4_prompt}, and examples of generated conversations before mapping the textual descriptions back to visual images are shown in Appx. \ref{app:conv_examples}. In the prompt, we also provide in-context examples to improve the generation quality. We collect the in-context examples through a human-in-the-loop refinement process, which is elaborated in \textsection\ref{sec:hitl}.
\vspace{-2mm}
\paragraph{Post-processing and Low-quality Filtering} To ensure data quality, we filter out conversations where there is a pair of input and output image descriptions with an edit distance higher than $0.1$. We also exclude conversations containing image descriptions not present in the provided image list and conversations containing formatting errors such as co-reference errors and invalid image tags.
\subsection{Human-in-the-loop Refinement \label{sec:hitl}}
In-context learning has been demonstrated to be crucial for enhancing the generation quality of LLMs \citep{brown2020language,wang-etal-2023-self-instruct}. Therefore, we also construct a seed set of high-quality in-context examples $\mathcal{S}$. The seed set $\mathcal{S}$ begins as an empty set and is iteratively updated with human feedback. In each iteration, we follow the steps detailed below:
\begin{enumerate}
    \item We employ the latest $\mathcal{S}$ and the template in Appx. \ref{app:gpt4_prompt}, and generate 100 new conversations using \ours (\textsection\ref{sec:method}).
    \vspace{-2mm}
    \item We manually analyze the generated conversations. Each conversation is assigned a quality label (``Excellent'', ``Satisfactory'', or ``Poor''). Besides, we label the visual abilities required for each conversation.The detailed annotation guideline for quality labels and visual abilities is outlined in Tab. \ref{tab:annotation} (Appx. \ref{app:annotation}).
    \vspace{-2mm}
    \item We add the generated conversations with ``Excellent'' or ``Satisfactory'' labels to $\mathcal{S}$.
\end{enumerate}
To ensure diversity across different conversations, we randomly sample three in-context examples from the seed set for each generation. We further require that at least one in-context example is labeled ``Excellent'' and the three examples must cover all four visual abilities. After three iterations, we fix the seed set and employ it to generate the remaining data. The percentage of ``Poor'' data annotated by humans declines from $30\%$ to $9\%$.

\section{\ours Data from GPT4 \label{sec:stats}}
We apply \ours to GPT4 and the Conceptual Captions (CC3M) dataset \citep{DBLP:conf/acl/SoricutDSG18, DBLP:conf/cvpr/ChangpinyoSDS21} as a case study. The details of the construction process can be found in Appx. \ref{app:data_implementation_details}. In this section, we present comprehensive analyses of the constructed dataset.

\begin{table}[t]
\centering
\small
\resizebox{0.85\linewidth}{!}{
\begin{tabular}{lc}
\toprule
\textbf{Statistics} &  \\\midrule
\# of conversations  & $25,629$ \\
Avg. \# turns in conversations & $3.36$ \\
Avg. \# images &   \\
\ \ \ \ in conversations &  $2.46$ \\
\ \ \ \ in instructions & $0.94$ \\
\ \ \ \ in responses & $1.52$ \\
Avg. \# words &  \\
\ \ \ \ in conversations & $285.90$ \\
\ \ \ \ in instructions & $78.66$ \\
\ \ \ \ in responses & $207.24$ \\
\bottomrule
\end{tabular}
}
\caption{Statistics of the dataset by applying \ours to GPT-4.\label{tab:stats}}
\end{table}

\begin{figure*}[t]
  \centering
  % First row
  \begin{subfigure}[b]{0.3\linewidth}
    \includegraphics[width=\textwidth]{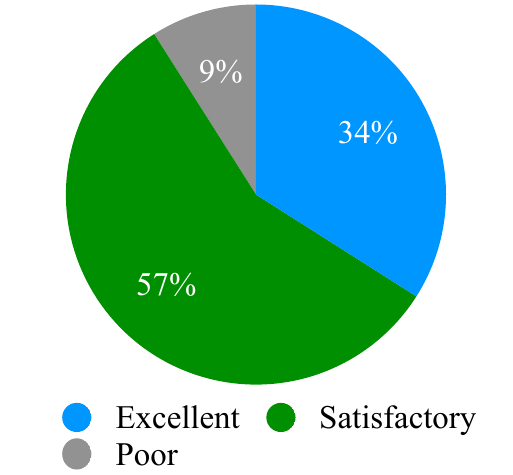}
    \caption{\label{fig:quality_dist}}
  \end{subfigure}
  % \hfill
  \begin{subfigure}[b]{0.33\linewidth}
    \includegraphics[width=\textwidth]{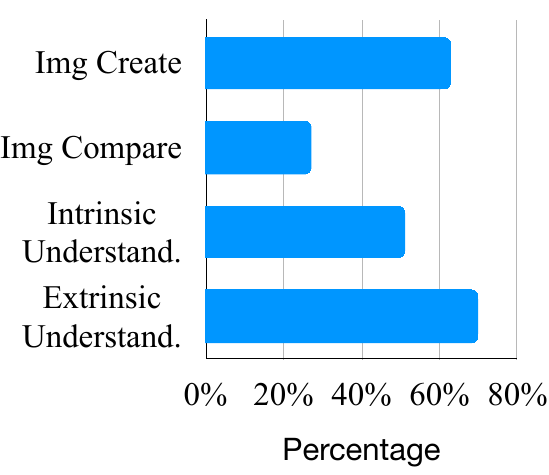}
    \caption{\label{fig:char_dist}}
  \end{subfigure}
  % \hfill
  \begin{subfigure}[b]{0.33\linewidth}
    \includegraphics[width=\textwidth]{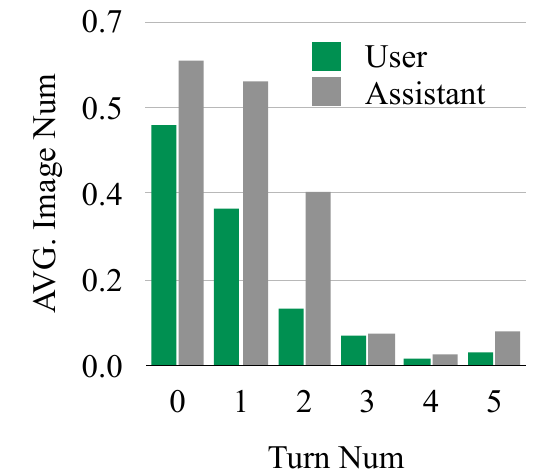}
    \caption{\label{fig:turn_image}}
    \end{subfigure}
  \caption{Statistics of data quality and diversity. The results in  Fig. \ref{fig:quality_dist} and \ref{fig:char_dist} are based on the human annotations on 100 randomly sampled conversations.}

\end{figure*}

\paragraph{Statistics}
As depicted in Tab. \ref{tab:stats}, our constructed dataset comprises $25,629$ conversations. The average number of turns per conversation is $3.36$ (each turn is defined as a pair of instruction and response). The mean number of images in each conversation is $2.46$. The length of each dialogue is about $285.90$.
\paragraph{Diversity\label{sec:div}} 
To understand the lexical and task diversity of our constructed data, we identify four types of required visual abilities: image creation, image comparison, intrinsic understanding, extrinsic understanding (See Tab. \ref{tab:annotation} for detailed definitions). The distribution is shown in Fig. \ref{fig:char_dist}. We observe that a significant portion of conversations in our dataset focuses on more insightful and informative tasks, such as extrinsic understanding and image comparison. For topic diversity, we display three randomly sampled clusters in Fig. \ref{fig:illustration}. The distribution of images across different turns is depicted in Fig. \ref{fig:turn_image}. We also compare the lexical diversity of our dataset and existing datasets in Tab. \ref{tab:text_div}. It can be seen that our datasets exhibit higher lexical diversity than existing datasets.
\paragraph{Quality}
To check the quality of the generated data, we randomly sample 100 conversations and perform an in-depth error analysis. The percentages of ``Excellent'', ``Good'', and ``Poor'' data are $57\%$, $34\%$, and $9\%$ respectively. Note that we label the whole conversation as ``Poor'' if any of its turns has a problem. We identify three major sources: image-caption mismatch, incoherence, and hallucination. Detailed analysis is in Appx. \ref{app:bad_example}.

\section{Augmenting LLMs with Visual I/O}
\subsection{Model}
\label{sec:model}
To support interleaved multimodal inputs and outputs, we supplement LLMs with visual input and output modules. Specifically, LLama2-Chat\footnote{\scriptsize{\url{https://huggingface.co/meta-llama/Llama-2-7b-chat-hf}}} \citep{touvron2023llama} is employed as the backbone LM. For visual input, we use the vision encoder from BLIP2 \citep{li2023blip}\footnote{\scriptsize{\url{https://huggingface.co/Salesforce/blip2-flan-t5-xxl}}}, followed by a pretrained Q-Former model \citep{li2023blip} that maps the features from the vision model into the embedding space of the LM. Inspired by GILL \citep{koh2023generating}, we attempt to learn a mapping from the output space of the LM to the input space of a stable diffusion (SD) model \citep{rombach2022high} (in this work, the embeddings produced by the text encoder of Stable Diffusion XL \citep{podell2023sdxl}). To this end, we explore three model variants in our preliminary experiments.
%The training examples of the MIM model follow the standard of Llama-Chat, as shown in Appx. \ref{app:llama_data_struct}. The model is trained to minimize the cross-entropy loss on the assistant’s turns, conditioned on the entire preceding conversation history.
\begin{table}[t]
\centering
\small
\resizebox{0.9\linewidth}{!}{
\begin{tabular}{lccc}
\toprule
\textbf{Dataset} & \textbf{Instruct} & \textbf{Response} & \textbf{Overall} \\\midrule
\textsc{LLaVA} & $1.56$ & $1.84$ & $1.70$ \\
\textsc{MiniGPT-4} & $0.00$ & $1.11$ & $0.89 $\\
\textsc{MultiInstruct} & $0.51$ & $1.69$& $0.51$ \\
\textsc{Platypus} & $0.98$ & $0.75$ & $0.78$ \\
\textsc{Shikra} & $0.89$ & $1.08$ & $0.87$ \\\midrule
\ours & $\mathbf{1.76}$ & $\mathbf{1.92}$ & $\mathbf{1.84}$ \\
\bottomrule
\end{tabular}}
\caption{Averaged diversity scores of roles in various datasets. Details of this analysis are in Appx. \ref{app:cmp_data}. \label{tab:text_div}
}
\end{table}
\begin{figure*}[t]
    \begin{center}
        \includegraphics[width=0.95\linewidth]{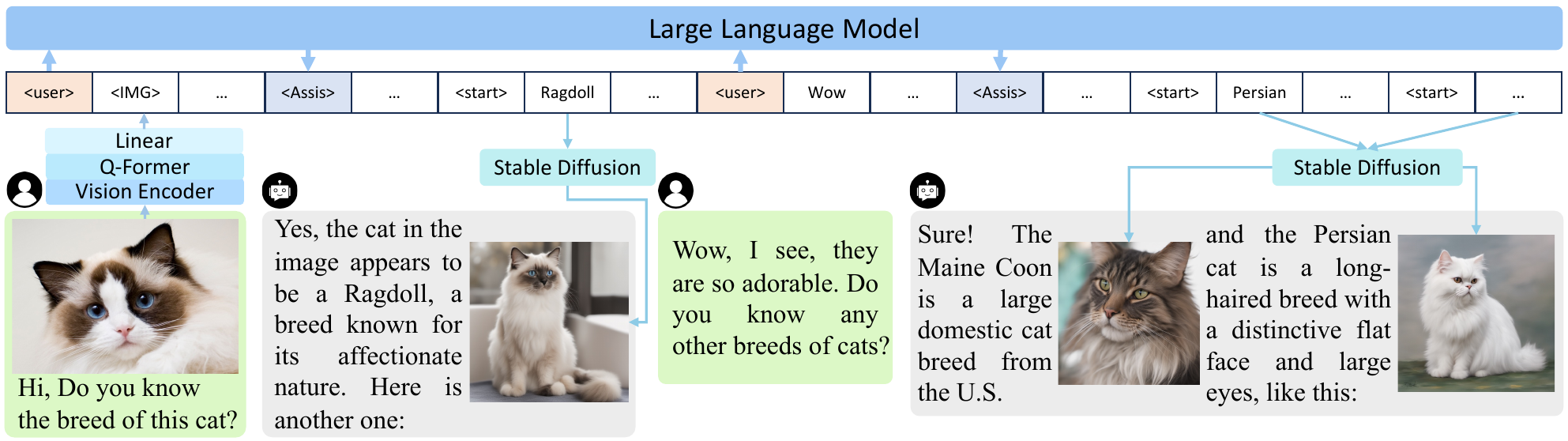}
    \end{center}
        \caption{The architecture of \ourmodel. It integrates a vision model, a language model, and a stable diffusion model. \ourmodel is able to process multi-turn interleaved multimodal inputs and outputs. Notably, the images generated by stable diffusion model will be fed back to the MIM model to enhance the image understanding. 
        \label{fig:arch}}
\end{figure*}

\noindent\textbf{Q-Former as Medium}. We add a special token \texttt{<IMG>} to the vocabulary of the LM, indicating that an image should be generated when it is emitted. We then use a Q-Former \citep{li2023blip} that takes all previous hidden states of the LM as input and outputs the SD embeddings.

\noindent\textbf{Q-Former with Prompt Tokens as Medium}. To further leverage the reasoning abilities of the LM, we incorporate a series of special tokens (\texttt{<img1>}, ..., \texttt{<IMG\{r\}>}), instead of a single token (\texttt{<IMG>}), to the LM. When \texttt{<img1>} is emitted, the generation of the special token sequence is enforced, serving as additional reasoning steps for predicting the forthcoming image. Subsequently, the Q-Former only accepts the hidden states of special tokens as input. 

\noindent\textbf{Language Description as Medium}. The previous two variants try to align the continuous hidden spaces of different models. An alternative is to use discrete language descriptions for information exchange, as depicted in Fig. \ref{fig:arch}. Specifically, we add two special tokens, \texttt{<start>} and \texttt{<end>}, and encode the generated text between these two tokens using the text encoder in the SD model.

The first two variants are optimized by minimizing the mean squared error (MSE) loss between the output and SD embeddings. For the third variant, we employ the standard cross-entropy loss. We empirically find that only the last method demonstrates satisfactory performance on \textbf{m}ulti-turn \textbf{i}nterleaved \textbf{m}ultimodal instruction-following, for which we name it \ourmodel. The key difference between MIM and GILL is that MIM uses discrete text as medium rather than continuous embeddings. See \textsection\ref{sec:exp_image_generation} for detailed comparisons.

\subsection{Training}
\label{sec:training}
Our training process consists of two stages, namely, the multimodal alignment stage and the multimodal instruction tuning stage.
\vspace{-2mm}
\paragraph{Multimodal Alignment}
The first stage aims to align the feature spaces of the vision model and the language model.
We utilize massive image-caption pairs for training, drawing from datasets such as Conceptual Captions \citep{DBLP:conf/cvpr/ChangpinyoSDS21, DBLP:conf/acl/SoricutDSG18} and SBU \citep{DBLP:conf/nips/OrdonezKB11}. During training, only the Q-Former connecting the vision and language models is optimized while other model components remain frozen.
\vspace{-2mm}
\paragraph{Multimodal Instruction Following}
The second stage further trains the joint model on multimodal instruction tuning data to improve its instruction-following capabilities. The Q-Former model and LLM are optimized in this stage. In addition to \ours, we also explore existing multimodal instruction data, including MultiInstruct \citep{xu-etal-2023-multiinstruct}, MiniGPT-4 \citep{zhu2023minigpt}, LLaVA \citep{liu2023llava}, and Shikra \citep{chen2023shikra}.

\section{Experiments\label{sec:exp}}
To verify the effectiveness of the proposed methods, we carry out quantitative evaluations against a set of recent baselines. Our quantitative evaluations are divided into three parts: textual response generation, image generation, and a holistic evaluation of multimodal instruction-following.

\subsection{\ourbench}
To facilitate comprehensive and dedicated evaluation for instruction-following in realistic scenarios, we construct a new dataset named \ourbench. \ourbench is initially generated through the automatic pipeline of \ours (\textsection\ref{sec:method}) and subsequently refined by human annotators. These annotators are tasked with discarding low-quality examples or rectifying amendable issues such as revising incoherent or hallucinated content. After a rigorous review, we establish an evaluation dataset comprising 278 conversations in total.
\begin{table*}[t]
\small
\centering
\resizebox{0.85\textwidth}{!}{

\begin{tabular}{lcccccc}
\toprule
\textbf{Methods} & \textbf{BLEU-2} & \textbf{BLEU-4} & \textbf{ROUGE-2} & \textbf{ROUGE-L} & \textbf{BERTScore} & \textbf{Diversity}  \\
\midrule
GILL \citep{koh2023generating} & 3.97 & 1.44 & 4.61 & 13.97 & 0.847 & 0.902 \\
\midrule
MultiInstruct \citep{xu-etal-2023-multiinstruct}\footnote{Since the MultiInstruct \citep{xu-etal-2023-multiinstruct} dataset is too large, we randomly sample 400 examples for each task in MultiInstruct, and the final amount of data is 22K.} & 7.16 & 2.27 & 3.16 & 10.60 & 0.830 & 0.654 \\
MiniGPT-4 \citep{zhu2023minigpt} & 9.24 & 3.29 & 6.77 & 17.56 & 0.858 & 0.658 \\
LLaVA \citep{liu2023llava} & 12.16 & 4.41 & 8.66 & 19.79 & 0.872  & 0.852 \\
Shikra \citep{chen2023shikra} & 10.37 & 3.83 & 7.79 & 18.63 & 0.864 & 0.722 \\
\midrule
\ours & \underline{24.45} & \underline{11.83} & \underline{15.45} & \underline{28.69} & \underline{0.891} & \textbf{0.927} \\
\textsc{Mix} & \textbf{27.64} & \textbf{14.49} & \textbf{17.90} & \textbf{31.22} & \textbf{0.896} & \underline{0.912} \\
\bottomrule
\end{tabular}%
}
\caption{
Evaluation results of textual response generation on \ourbench.}
\label{tab:multimodal_instruciton_following}
\end{table*}
\subsection{Textual Response Generation\label{sec:exp_text}} 

% A crucial capability of multimodal LLMs is text generation within a multimodal context, which includes tasks such as describing images for users and answering questions based on images.

\paragraph{Setup} We consider each assistant turn of each conversation in \ourbench as a test point. All its preceding context is treated as input (which may contain interleaved images and text), and the goal is to generate a coherent and helpful response. We measure the response quality using a set of reference-based evaluation metrics such as BLEU \citep{DBLP:conf/acl/PapineniRWZ02}, ROUGE \citep{lin-2004-rouge}, and BERTScore \citep{DBLP:conf/iclr/ZhangKWWA20}. We also report the Diversity \citep{su2022a} scores of the generated responses. For simplicity, we replace any image in the responses with a special token \texttt{<image>}.

For a fair comparison, we compare different \ourmodel models trained on different datasets \citep{xu-etal-2023-multiinstruct,zhu2023minigpt,liu2023llava,chen2023shikra}\footnote{The original papers of these datasets used distinct model architectures such as different pretrained language models. Nevertheless, all of them do not support image generation.} and GILL \citep{koh2023generating}\footnote{For a fair comparison, we replicate GILL using the same image-captioning data to train by our models.}. The implementation details are shown in Appx. \ref{app:model_implementation_details}.
\begin{table}
  \centering
  \resizebox{0.99\linewidth}{!}{%
  \begin{tabular}{lcccc}
    \toprule
    \multirow{2}{*}{\textbf{Training Dataset}} &   \multicolumn{2}{c}{\textbf{MME}} &  \multicolumn{1}{c}{\textbf{MMBench} }  &  \rev{\textbf{MM-Vet}}   \\
    \cmidrule(r){2-3} 
    
    & Perception & Cognition & Overall & Overall \\
    \midrule
    MultiInstruct \citeyearpar{xu-etal-2023-multiinstruct} & \textbf{1099.16} & \textbf{302.50} & 31.54 & \rev{17.2} \\
    MiniGPT-4 \citeyearpar{zhu2023minigpt} & 0.00 & 0.00 & 31.87 & \rev{9.8}\\
    LLaVA \citeyearpar{liu2023llava} & 683.28 & \underline{267.86} &  \underline{42.10} & \underline{23.4} \\
    Shikra \citeyearpar{chen2023shikra} & 166.87 & 2.86 & 41.10 & \rev{19.9}\\
    \midrule
    \ours & 549.00 & 226.43 & 22.64 & \rev{19.4} \\
    \textsc{Mix} & \underline{1023.33} & 255.00 & \textbf{44.94} & \rev{\textbf{23.9}} \\
    \bottomrule
  \end{tabular}
  }
  \caption{Evaluation Results on MME \citep{Fu2023MMEAC}, MMBench \citep{liu2023mmbench}, \rev{MM-Vet \citep{yu2023mm}}.}
  \label{tab:mme_mmbench}
\end{table}
\paragraph{Results} As shown in Tab. \ref{tab:multimodal_instruciton_following}, the \ourmodel model trained on \ours outperforms all other baselines by wide margins across all evaluation metrics. The results suggest that more realistic and diverse training data such as \ours is necessary for tackling open-world tasks, which cannot be well-supported by existing template-based and VQA-like datasets. Nevertheless, we also find that the performance can be further improved when combining different datasets (i.e., the mixture of MultiInstruct, MiniGPT-4, LLaVA, Shikra, and \ours, denoted by \textsc{Mix}), indicating that there is a complementary relationship between \ours and existing datasets.

\paragraph{Additional Results on Conventional Benchmarks} Finally, we report the results on three popular multimodal benchmarks, MME \citep{Fu2023MMEAC}, MMBench \citep{liu2023mmbench}, and MM-Vet \citep{yu2023mm}. 
As shown in Tab. \ref{tab:mme_mmbench}, \ours gets a relatively lower score than other datasets. The reason stems from the intrinsic difference between \ours and the two benchmarks. \ours focuses more on realistic instructions (i.e., free-form text generation based on arbitrarily interleaved images and text across diverse user-assistant conversations). In contrast, MME, MMBench and MM-Vet focus more on VQA questions (e.g., who is this person, what is the color of the object), which are more similar to the data in MultiInstruct, LLaVA, and Shikra. For example, the model trained on MultiInstruct achieves the \textit{best} performance on MME, though it displays the \textit{worst} performance in open-world scenarios in Tab. \ref{tab:multimodal_instruciton_following}. Another interesting observation is that the mix of all datasets (\textsc{Mix}) attains the best overall performance on MMBench and MM-Vet, indicating that different datasets are complementary. This again confirms that the capabilities that \ours can bring are almost orthogonal to existing multimodal instruction-following datasets.
\subsection{Image Generation}
\label{sec:exp_image_generation}
\begin{table*}[t]
  \centering
  \resizebox{0.90\linewidth}{!}{%
  \begin{tabular}{lccccccccc}
    \toprule
    \multirow{2}{*}{\textbf{Model}} &&   \multicolumn{3}{c}{\textbf{CLIP Similarity} ($\uparrow$)} &&  \multicolumn{3}{c}{\textbf{LPIPS} ($\downarrow$)}   & \rev{\textbf{ FID ($\downarrow$)}}  \\
    \cmidrule(r){3-5} \cmidrule(r){7-9} \cmidrule(r){10-10} 
        && Turn-1  & Turn-2 &  Turn-3  &&  Turn-1  & Turn-2 &  Turn-3 & \rev{ All}\\
    \midrule
    SDXL \citep{podell2023sdxl} && 0.612 & 0.599 & 0.608 && \textbf{0.712} & 0.735 & 0.735 & \rev{ 144.76}
 \\
    GILL \citep{koh2023generating} && 0.569 & 0.550 & 0.530 && \textbf{0.712} & 0.734 & 0.742 & \rev{ 158.64} \\\midrule
    Q-Former as Medium && 0.558 & 0.568 & 0.592 && 0.717 & 0.728 & 0.729 & \rev{ 155.01} \\
    Q-Former with Prompt Tokens as Medium && 0.566 & 0.571 & 0.606 && 0.718 & 0.727 & 0.732 & \rev{ 152.23} \\
    \ourmodel (Language Description as Medium) && \textbf{0.640} & \textbf{0.645} & \textbf{0.673} && \textbf{0.712} & \textbf{0.720} & \textbf{0.726}  & \rev{{\textbf{139.46}}}
\\
    \bottomrule
  \end{tabular}
  }
  \caption{Evaluation results of image generation on \ourbench.}
  \label{tab:contextual_image_generation}
\end{table*}
\paragraph{Setup}
The models trained on existing datasets, i.e., the baselines in \textsection\ref{sec:exp_text} except for GILL, are incapable of generating images. To showcase the image generation capabilities of our model, we compare it with SDXL \citep{podell2023sdxl} and GILL \citep{koh2023generating}. In addition, we present the results of the two model variants described in \textsection\ref{sec:model}, namely, \textbf{Q-former as Medium} and \textbf{Q-former with Prompt Tokens as Medium}.

We take each image from the assistant in \ourbench as a test point. All its preceding context is taken as input, and the models are enforced to output an image. We take the original images in \ourbench as references. Following \citet{koh2023generating}, we evaluate image generation with two reference-based metrics: (1) \textbf{CLIP Similarity}. We use the CLIP vision encoder to produce image representations and compute the cosine similarity between generated images and reference images. A higher score means better semantic similarity. (2) \textbf{Learned Perceptual Image Path Similarity (LPIPS)}. LPIPS \citep{DBLP:conf/cvpr/ZhangIESW18} measures the distance between generated images and reference images. A lower score means that images are more similar in perceptual space. \rev{(3) \textbf{Frechet Inception Distance} (FID). FID measures the distributional difference between the generated images and reference images. A lower score indicates better resemblance to reference images.}

\paragraph{Results}
To gain further insights into the multi-turn instruction-following abilities, we group different test points by the number of previous conversation turns. The results are shown in Tab. \ref{tab:contextual_image_generation}. As seen, \ourmodel generally achieves better performance than SDXL and GILL across different turns and evaluation metrics. Importantly, the performance gaps are enlarged as the number of turns increases. This indicates that our model exhibits a better understanding ability of multi-turn conversations. Compared to the two model variants, \ourmodel is substantially better. Our case study reveals that the disparity stems from the \textit{one-to-many} nature of image generation in real-world conversations. Unlike generating images for explicit descriptions, there can exist numerous distinct images for a given conversation context. Operating in the hidden space may inadvertently average all possibilities, resulting in ambiguous or noisy images. However, \ourmodel mitigates the \textit{one-to-many} issue by taking full advantage of the autoregressive generation of language models for decision-making.

\subsection{Holistic Evaluation}
In addition to the above automatic evaluation, we also conduct a holistic evaluation of instruction-following abilities through human annotation.

\begin{table}
\small
\centering
\resizebox{0.95\linewidth}{!}{
\centering
    \begin{tabular}{lcc|ccc}
    \toprule
    \textbf{Methods} & \textbf{Overall} & \textbf{Acc. rate} & \textbf{IR} & \textbf{CU} &\textbf{RI}  \\
    \midrule
    GILL & $1.71$ & $19\%$  & - & - & -\\
    LLaVA & $2.93$  & $70\%$  & 3.59    & 3.56                            &3.78 \\\midrule
    \ourmodel  & $3.39$ & $89\%$   & 3.99                & 3.82                            & 3.72  \\\bottomrule
    \end{tabular}%
}
\caption{Human evaluation results. Acc. rate denotes the percentage of cases with an average score $\geq 3$. IR, CU, and RI represent intent recognition, context understanding and response informativeness, respectively. \label{tab:human_eval}}
\end{table}

\paragraph{Setup}
We randomly sample 100 contexts from \ourbench and evaluate the responses generated by \ourmodel and two representative baselines, LLaVA \citep{liu2023llava} and GILL \citep{koh2023generating}. We instruct three human annotators to score the quality of each generated response on a Likert scale from 1 to 4 (major error, minor error, acceptable, and excellent). The details of evaluation guideline are in Appx. \ref{app:human_evaluation}. For inter-annotator agreement, the Krippendorff’s $\alpha=0.75$, which indicates a high consistency across annotators.

\paragraph{Results} As depicted in Table \ref{tab:human_eval}, \ourmodel achieves substantially higher human scores than GILL and LLaVA, in terms of both average score and acceptance rate, indicating its remarkable superiority in open-world multimodal conversations. To further elucidate where the derived dataset and training helps, we ask human annotators to evaluate the quality of the generated responses across three fine-grained dimensions: intent recognition, context understanding and response informativeness. As seen, \ourmodel is more adept at following human instructions and leveraging the multimodal context, while the informativeness scores of models trained on \ours and LLaVA remain comparable.

% \begin{table*}
%   \centering
%   \resizebox{1.0\linewidth}{!}{%
%   \begin{tabular}{lcccccccccccc}
%     \toprule
%     &&   \multicolumn{2}{c}{\textbf{MME}} &&  \multicolumn{7}{c}{\textbf{MMBench} }  &  \rev{\textbf{MM-Vet}}   \\
%     \cmidrule(r){3-4} \cmidrule(r){6-12} \cmidrule(r){13-13}
%     \textbf{Training Dataset} && Perception & Cognition && LR & AR & RR & FP-S & FP-C & CP & Overall & \rev{-} \\
%     \midrule
%     MultiInstruct \citeyearpar{xu-etal-2023-multiinstruct} && \textbf{1099.16} & \textbf{302.50} && 11.93 & 39.79 & 28.64 & 28.75 & 23.20 & 41.91 & 31.54 &  \rev{17.2} \\
%     MiniGPT-4 \citeyearpar{zhu2023minigpt} && 0.00 & 0.00 && \textbf{14.20} & 50.52 & 17.37 & 32.75 & 15.20 & 41.70 & 31.87 & \rev{9.8}\\
%     LLaVA \citeyearpar{liu2023llava} && 683.28 & 267.86 && 7.95 & 55.71 & 31.46 & 42.50 & 31.60 & \textbf{56.60} & 42.10 & \rev{23.4} \\
%     Shikra \citeyearpar{chen2023shikra} && 166.87 & 2.86 && 18.18 & \textbf{64.01} & 22.54 & 39.75 & 31.20 & 50.43 & 41.10 & \rev{19.9}\\
%     \ours && 549.00 & 226.43 && 11.93 & 36.33 & 6.57 & 23.25 & 6.00 & 33.83 & 22.64 & \rev{19.4} \\
%     Mix && 1023.33 & 255.00 && 13.64 & 56.75 & \textbf{37.09} & \textbf{43.50} & \textbf{42.80} & 55.32 & \textbf{44.94} & \rev{\textbf{23.9}} \\
%     \bottomrule
%   \end{tabular}
%   }
%   \caption{Results on MME \citep{Fu2023MMEAC}, MMBench \citep{liu2023mmbench}, \rev{MM-Vet \citep{yu2023mm}}.}
%   \label{tab:mme_mmbench}
% \end{table*}

\section{Conclusion}
In conclusion, the introduction of the \ours framework has opened new doors for enhancing large language models with multi-turn interleaved multimodal instruction-following capabilities. By requiring only image-caption pairs, our approach significantly reduces the need for high-quality exemplar data, making it a more accessible and scalable solution for various real-world tasks. The \ourmodel architecture seamlessly integrates image encoder and decoder models, enabling the model to effectively handle interleaved image-text inputs and outputs. Comprehensive quantitative and qualitative experiments demonstrate the remarkable performance of \ourmodel, trained on \ours, when compared to recent baselines in  open-world multimodal conversations.

\section*{Limitations}
Despite the amazing instruction-following capabilities demonstrated by our models trained on \ours, there are many avenues for further improvement.
\begin{enumerate}
    \item Our best-performing models use textual descriptions as a medium for image generation. Nevertheless, it is impossible to describe everything in an image through words. Thus, our current models have difficulties completing tasks such as precise image editing. Investigating fine-grained multimodal feature combinations may enhance performance in these areas.
    \item The multimodal capabilities of \ours can be further enriched by incorporating additional modalities. For example, future work may use audio-text pairs for data synthesis.
    \item Similar to existing multimodal language models, models trained on \ours exhibit some common deficiencies such as hallucination, toxicity, and stereotypes.
\end{enumerate}

\bibliography{custom}

\clearpage

\appendix

\section{Demonstrations\label{app:demonstration}}
The four high-level characteristics of the \ours dataset equips \ourmodel with a variety of capabilities. We demonstrate those capabilities with concrete user cases.

\begin{figure*}
    \begin{center}
        \includegraphics[width=0.8\linewidth]{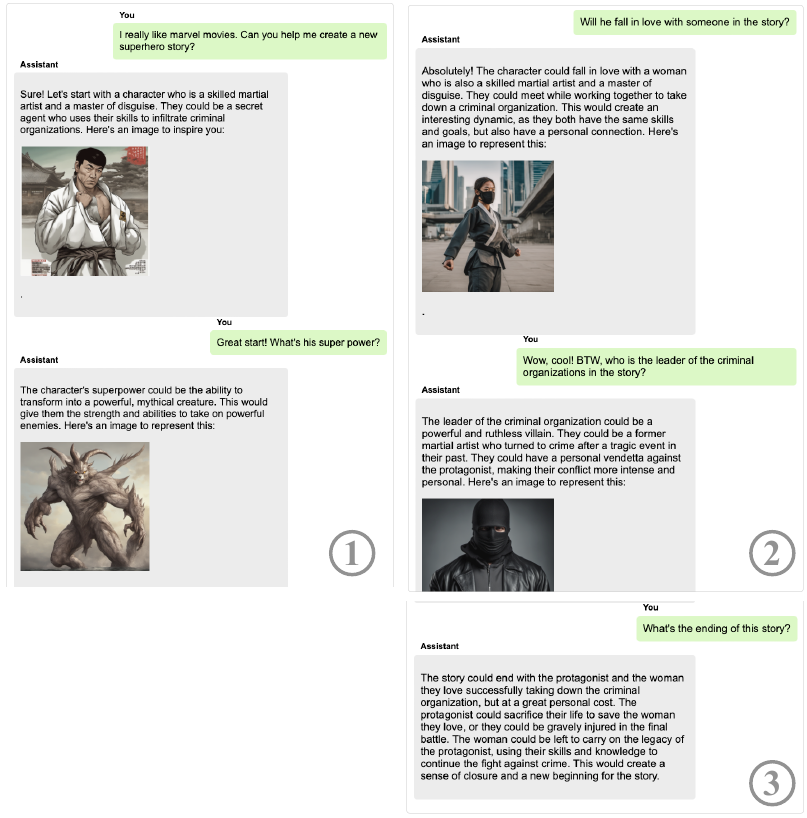}
    \end{center}
        \caption{Generation of a long story with interleaved text and images.
        \label{fig:long_story}}
\end{figure*}

\begin{figure*}
  \centering
  % First row
  \begin{subfigure}[b]{0.47\linewidth}
    \includegraphics[width=\textwidth]{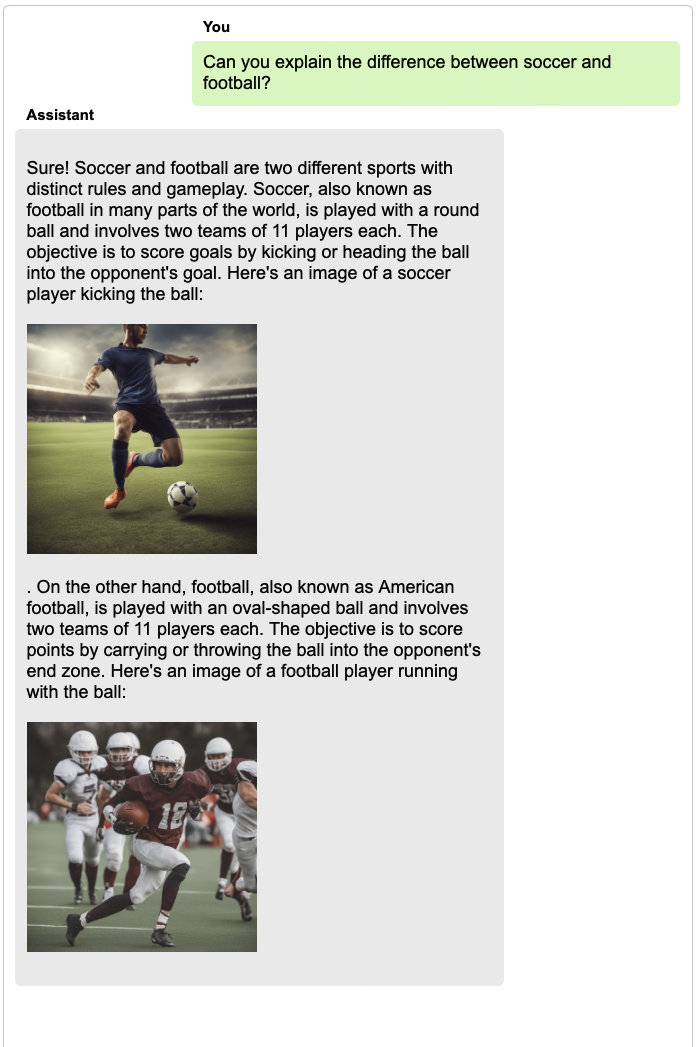}
    \caption{Explaining concepts with multiple images.\label{fig:creat_1}}
  \end{subfigure}
  \begin{subfigure}[b]{0.48\linewidth}
    \includegraphics[width=\textwidth]{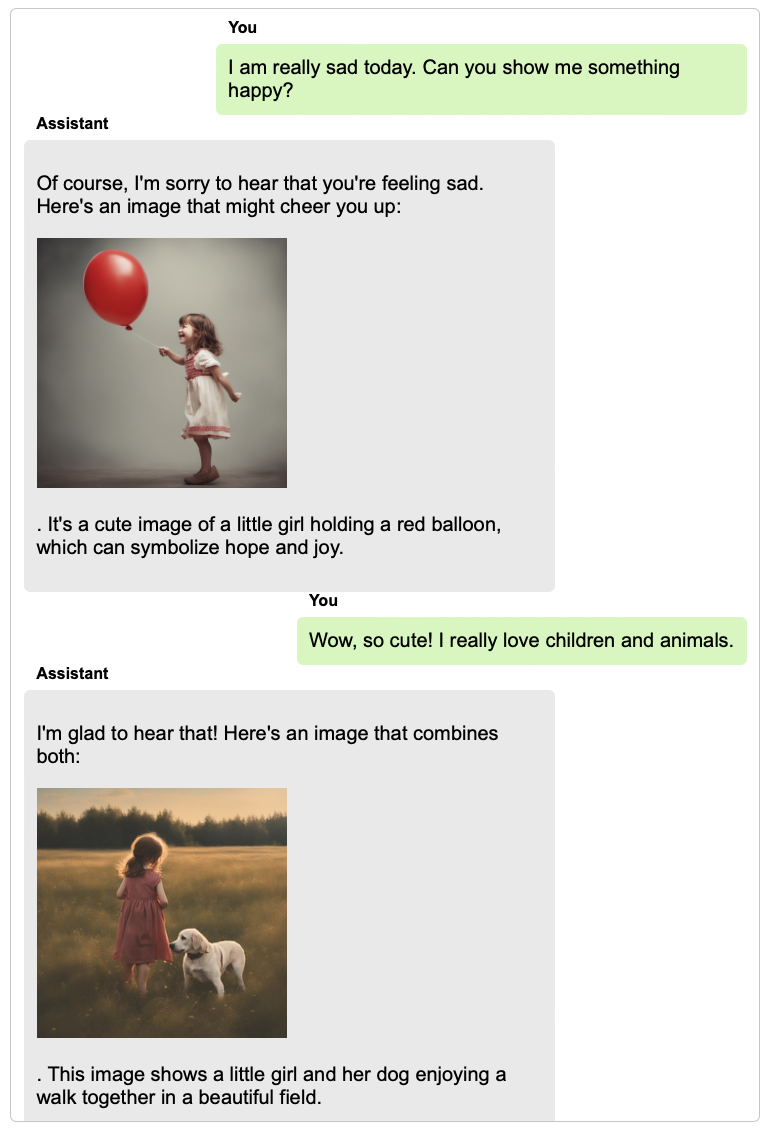}
    \caption{Creating images with correct emotions.\label{fig:creat_2}}
  \end{subfigure}

  % \begin{subfigure}[b]{0.48\linewidth}
  %   \includegraphics[width=\textwidth]{./figs/ex8.png}
  %   \caption{Suggestions with images. \label{fig:creat_3}}
  
  % \end{subfigure}
  \begin{subfigure}[b]{0.48\linewidth}
    \includegraphics[width=\textwidth]{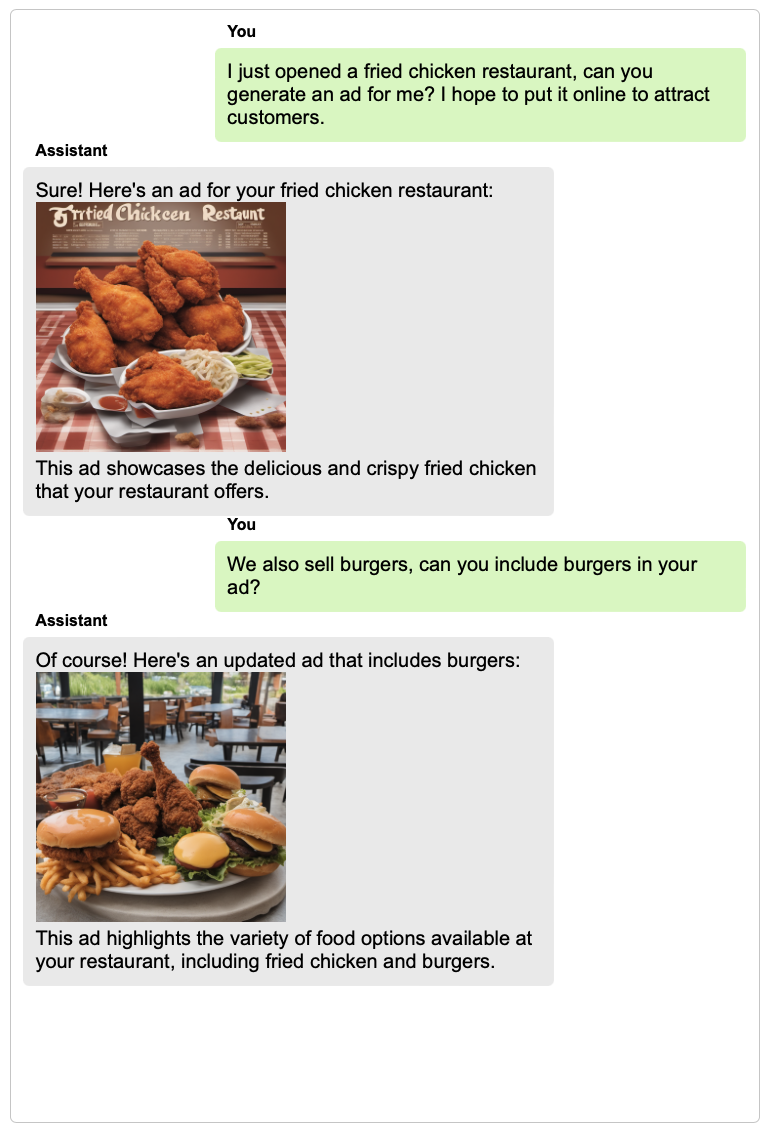}
    \caption{Editing images based on context.\label{fig:creat_3}}
  \end{subfigure}
  \begin{subfigure}[b]{0.48\linewidth}
    \includegraphics[width=\textwidth]{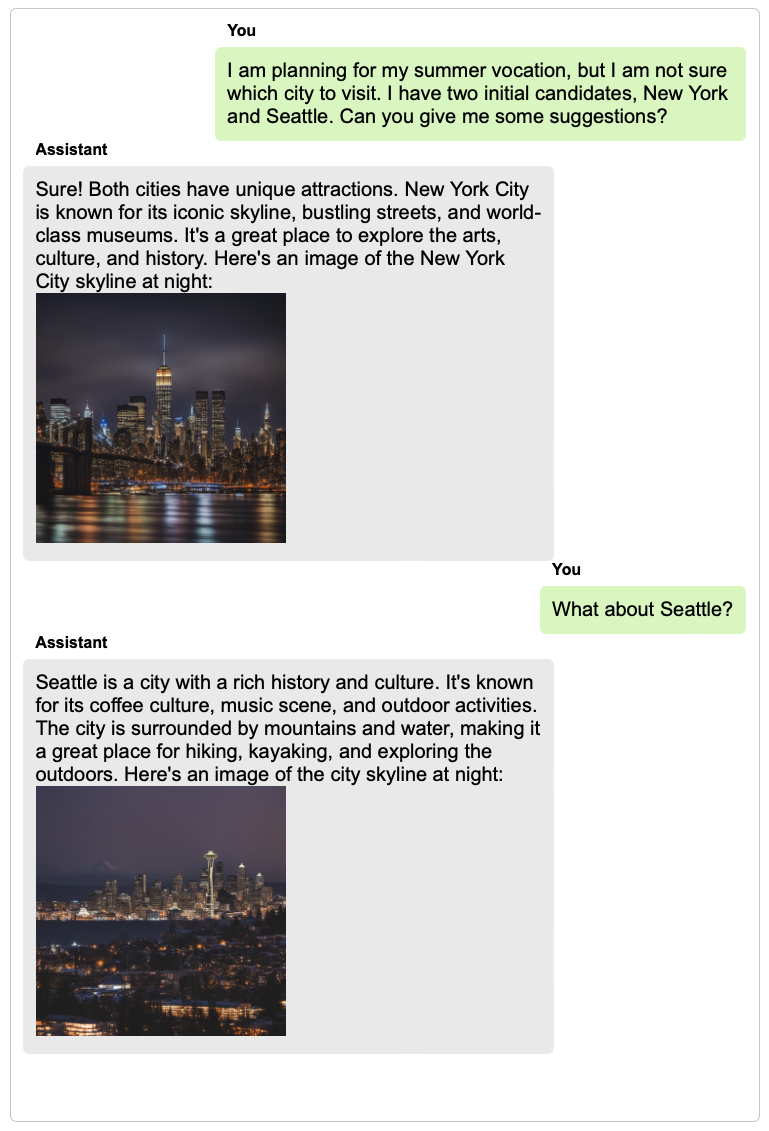}
    \caption{Creating images based on context.\label{fig:creat_4}}
  \end{subfigure}
  
  \caption{User cases of creating images. \label{fig:creat}}
\end{figure*}

\paragraph{Image Creation} One core innovation of \ours is that it enables the model to create images based on the conversation context without explicit dictations from the users. This characteristic is extremely useful for open-world scenarios, because in many cases people may just have an implicit intention and have no clear thoughts about what the images should be. We observe that our model can explain concepts and ideas for users with vivid images (Figure \ref{fig:creat_1}), creating images with correct emotions (Figure \ref{fig:creat_2}), and editing images based on the whole context (Figure \ref{fig:creat_3} and \ref{fig:creat_4}). Furthermore, as shown in Figure \ref{fig:long_story}, we discover that our model is proficient in generating long stories featuring interleaved text and images while maintaining exceptional coherence.

\begin{figure*}
  \centering
  % First row
  \begin{subfigure}[b]{0.47\linewidth}
    \includegraphics[width=\textwidth]{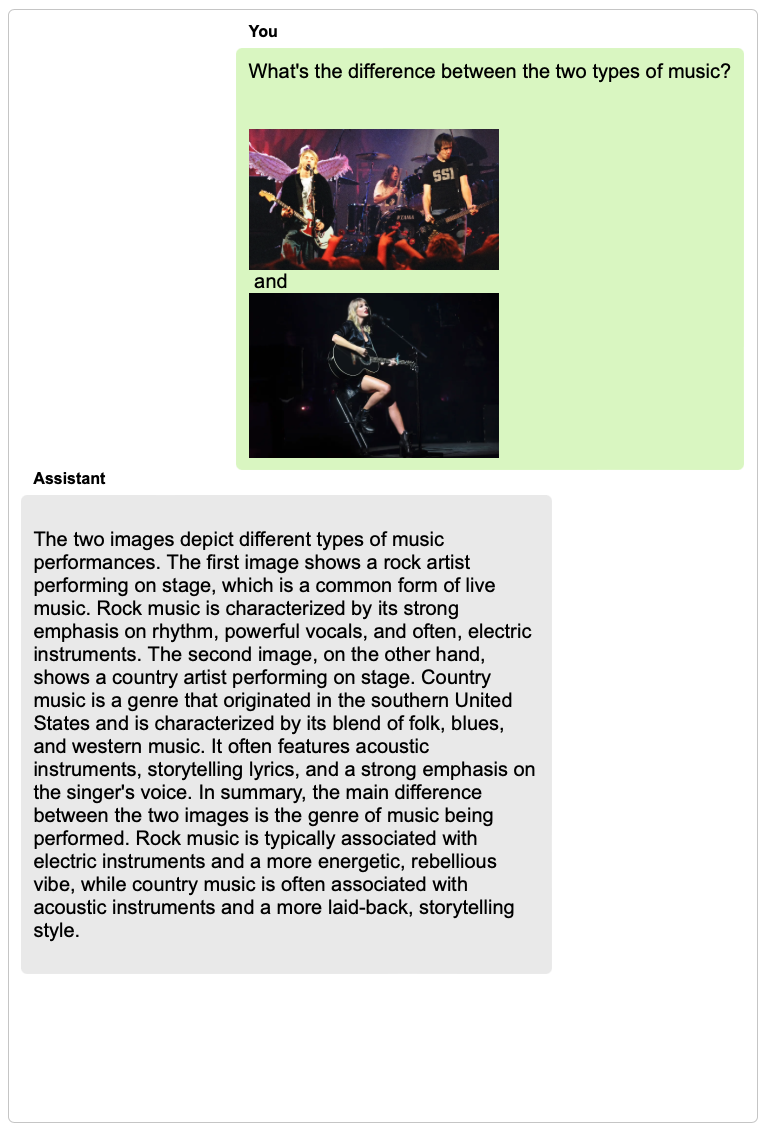}
    \caption{Comparing music styles.\label{fig:compare_1}}
  \end{subfigure}
  \begin{subfigure}[b]{0.47\linewidth}
    \includegraphics[width=\textwidth]{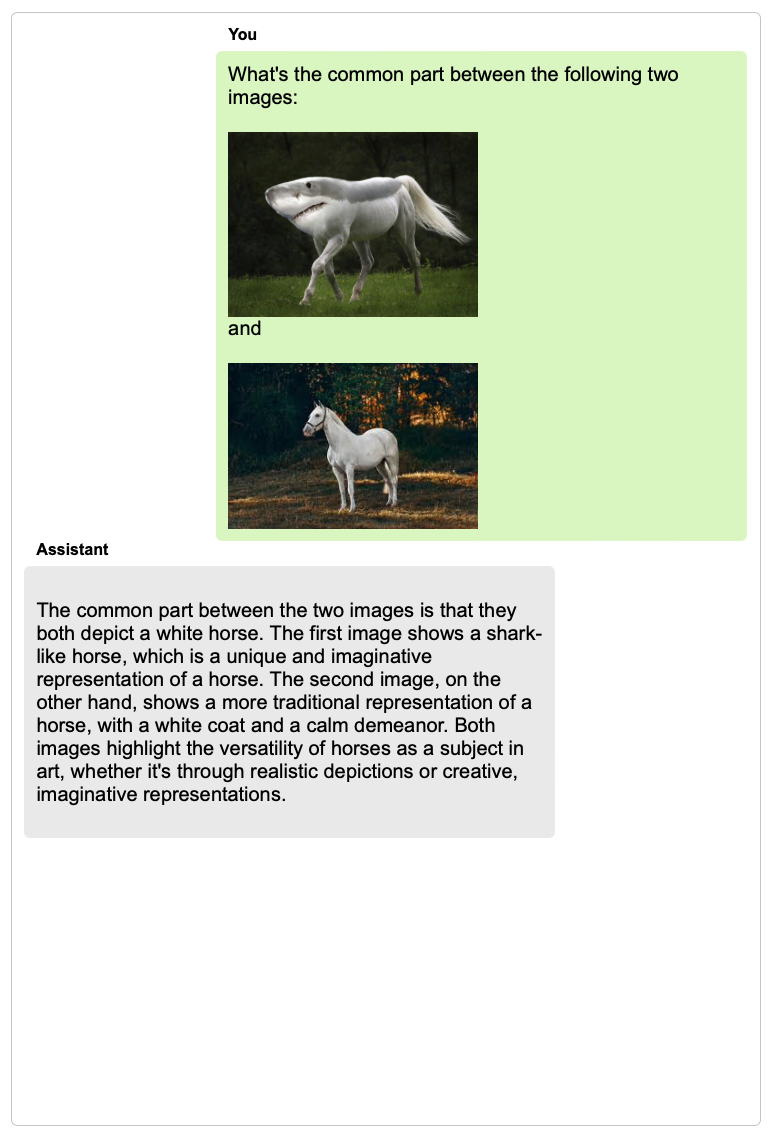}
    \caption{Relating images.\label{fig:compare_2}}

  \end{subfigure}

  \begin{subfigure}[b]{0.47\linewidth}
    \includegraphics[width=\textwidth]{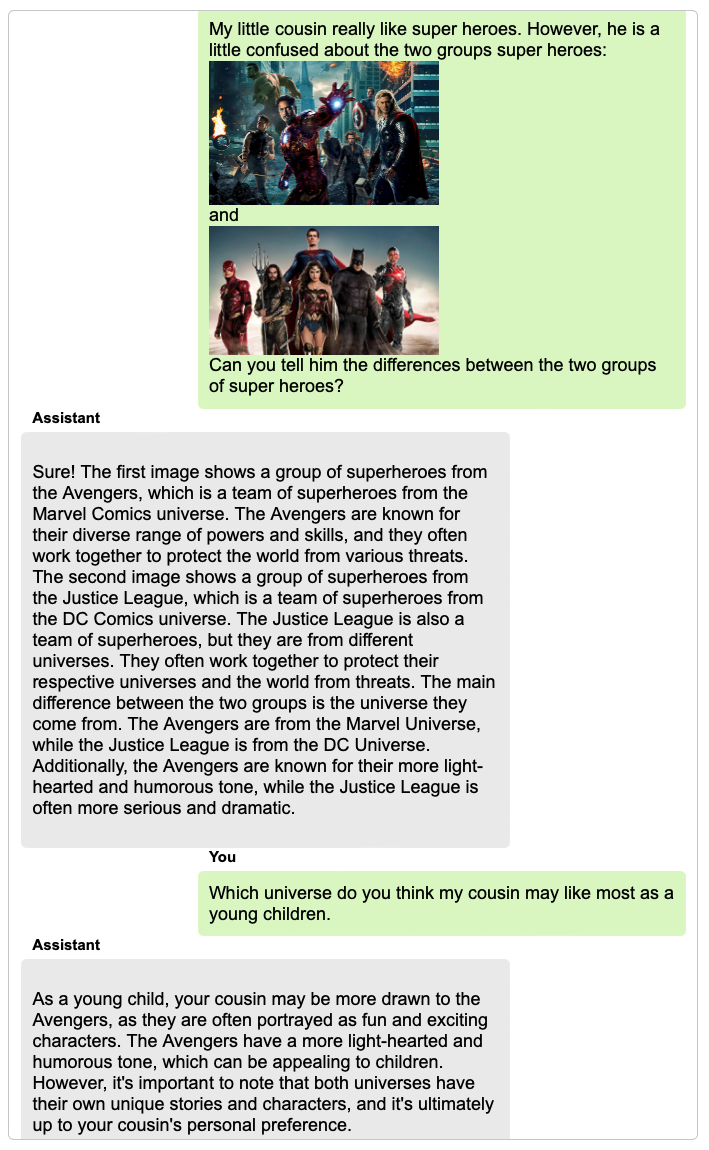}
    \caption{Comparing movies. \label{fig:compare_3}}
  \end{subfigure}
  \begin{subfigure}[b]{0.48\linewidth}
    \includegraphics[width=\textwidth]{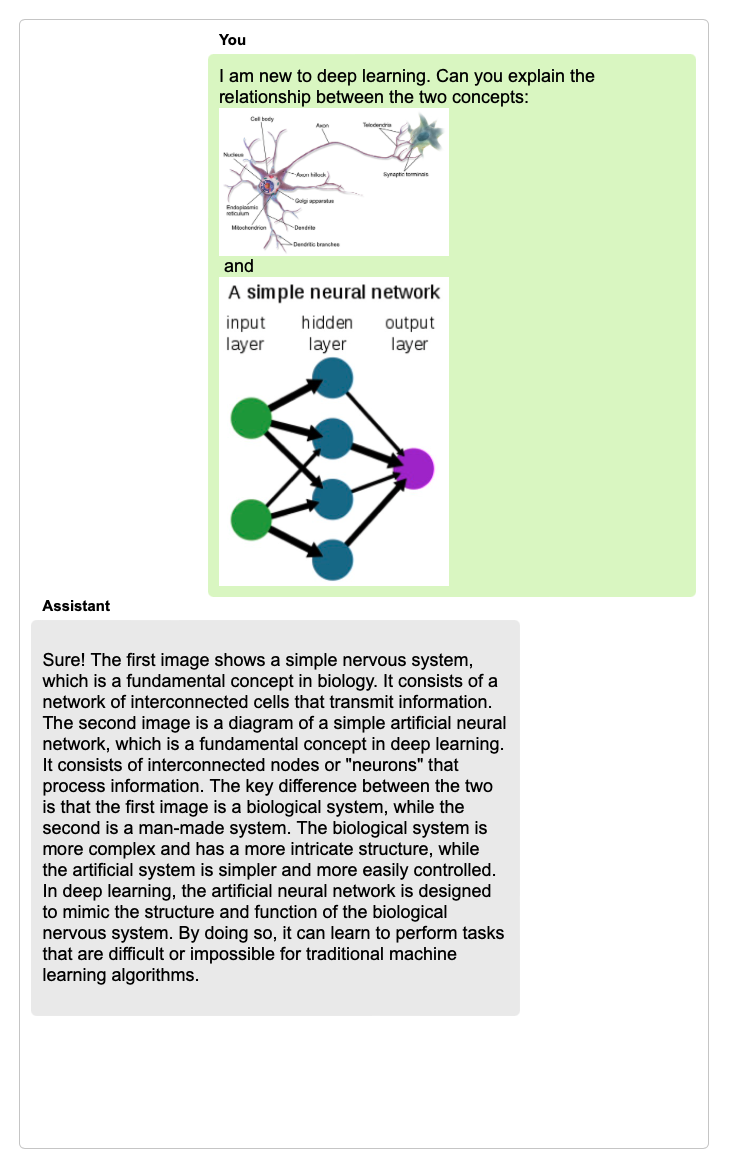}
    \caption{Comparing different concepts. \label{fig:compare_4}}
  \end{subfigure}

  \caption{User cases of comparing images. \label{fig:compare}}
  \label{fig:exampleImages}
\end{figure*}

\paragraph{Image Comparison} Another interesting feature of \ours is that it can compare or relate the information in multiple images. For example, our model can correctly explain the different and common parts in multiple images in Figure \ref{fig:compare}.

\begin{figure*}
  \centering
  % First row
  \begin{subfigure}[b]{0.45\linewidth}
    \includegraphics[width=\textwidth]{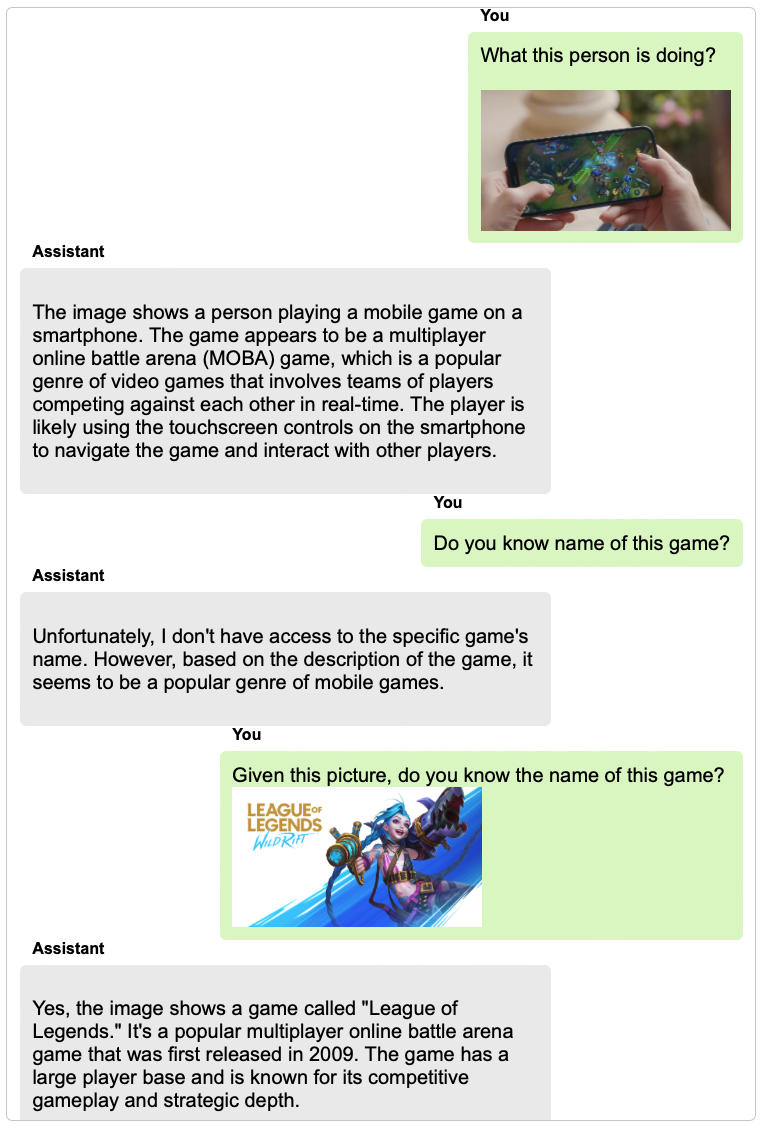}
    \caption{\label{fig:understand_1}}
  \end{subfigure}
  \begin{subfigure}[b]{0.45\linewidth}
    \includegraphics[width=\textwidth]{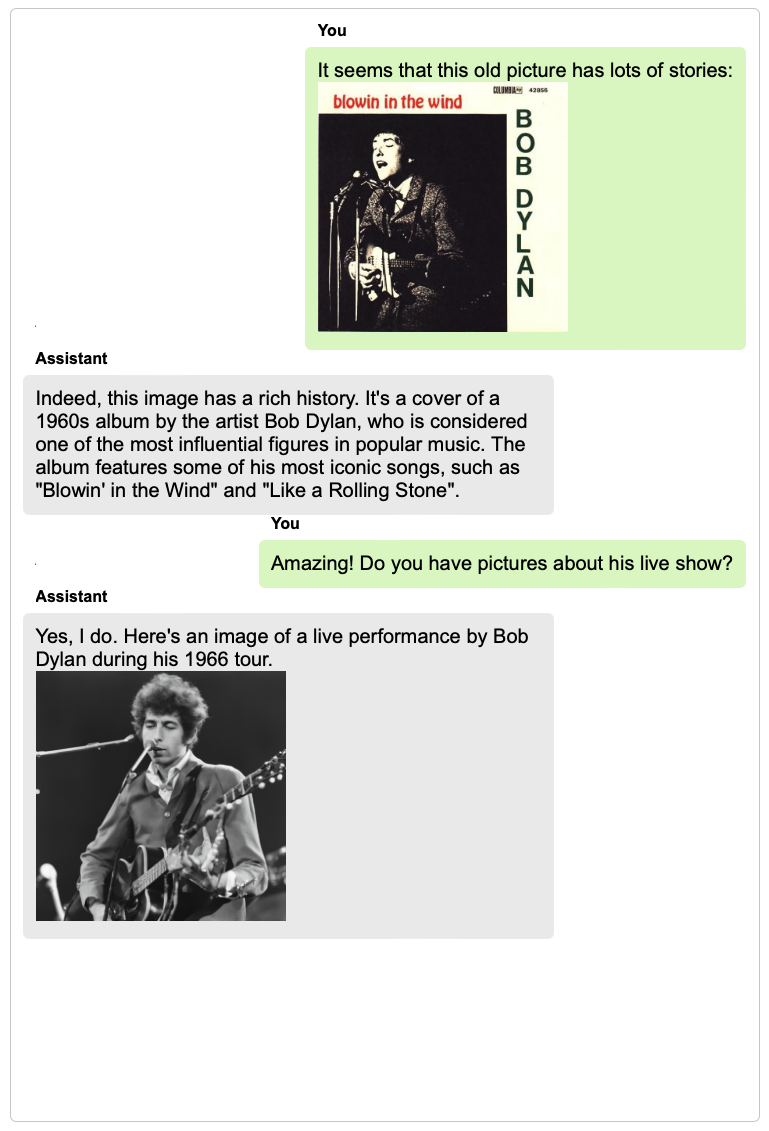}
    \caption{\label{fig:understand_2}}
  \end{subfigure}

  \begin{subfigure}[b]{0.49\linewidth}
    \includegraphics[width=\textwidth]{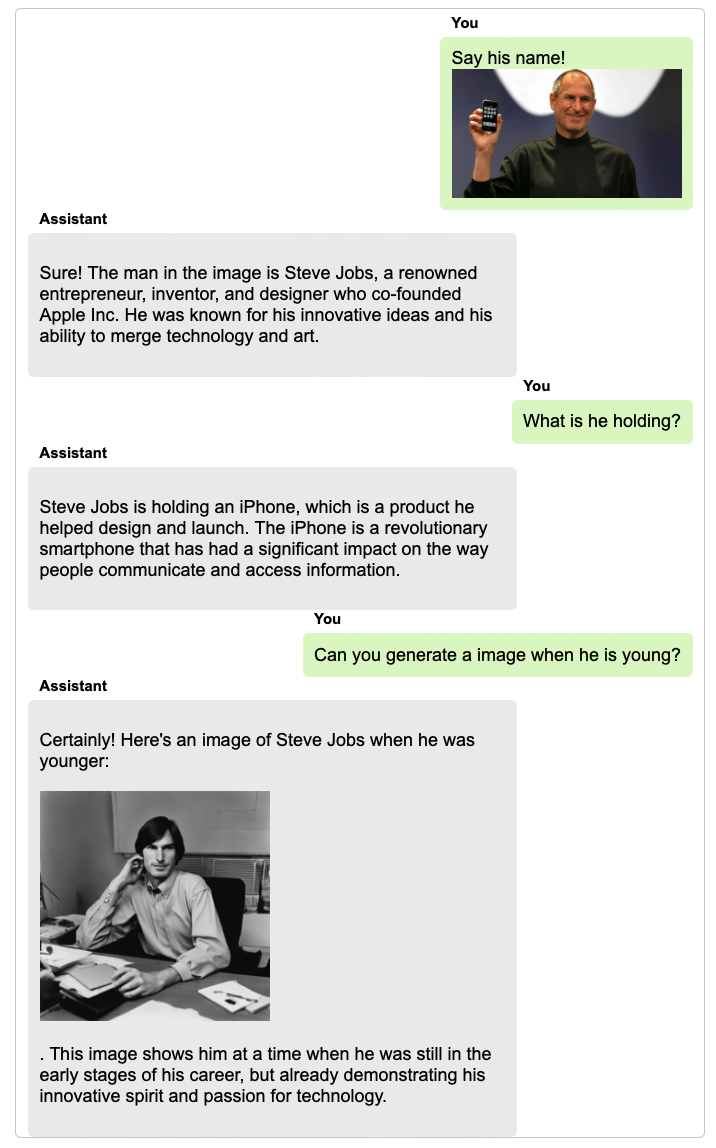}
    \caption{\label{fig:understand_3}}
  \end{subfigure}
  \begin{subfigure}[b]{0.42\linewidth}
    \includegraphics[width=\textwidth]{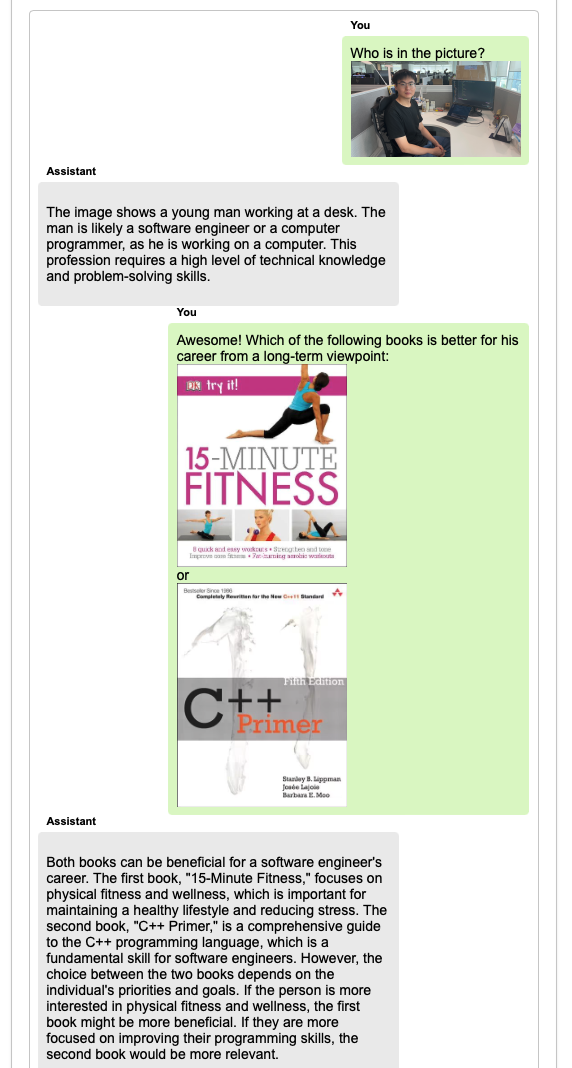}
    \caption{\label{fig:understand_4}}
  \end{subfigure}

  \caption{User cases of understanding both intrinsic \& extrinsic information in the images. \label{fig:understand}}
\end{figure*}

\paragraph{Intrinsic \& Extrinsic Image Understanding} The model trained on \ours can understand the content in images precisely in a multi-turn conversation. In all the three sub-figures of Figure \ref{fig:understand}, the model precisely follows the human instructions and explains the details of the images to users. Moreover, \ours also enables the model to explore the meaning of an image beyond the symbols in it. For example, the model also explains the the influence of Bob Dylan's album in Figure \ref{fig:understand_2} and the impact of iPhone in Figure \ref{fig:understand_3}.

\section{Prompt of \ours\label{app:gpt4_prompt}}
The prompt is shown in Fig. \ref{fig:app_gpt4_prompt}.

\begin{figure*}
\centering
\resizebox{0.9\textwidth}{!}{
\begin{AIbox}{GPT-4 Prompt}
Please construct a dialogue between a human and a helpful, honest and harmless assistant. The dialogue contains interleaved text and images. Each image is represented by {\color{red}$<$imgX$>$ DESCRIPTION $<$/imgX$>$}, where {\color{blue} DESCRIPTION} is a textual description of the image and {\color{blue} X} is an index of the image. Please do not assume any further visual information beyond the description.\\

The constructed dialogues must and can only contain the following input images:\\
{\color{forestgreen}$<$img0$>$} museum - the 1st nuclear submarine {\color{forestgreen}$<$/img0$>$}\\
{\color{forestgreen}$<$img1$>$} response to the production of heavy {\color{forestgreen}$<$/img1$>$}\\

{\bf Characteristics about the assistant:}\\
1. The assistant is trained to understand text, images, and their combinations.\\
2. The assistant can reply to the human with images and/or text.\\
3. The assistant has exceptional world knowledge and commonsense reasoning capabilities.\\
4. The assistant does not have access to the Internet or any other external tools.\\
5. If the assistant is asked to create an image, it can only show the image in the provided image list.\\
6. Please do not copy the images appearing in the dialogue. The assistant should refer to the previously mentioned image by natural language. \\

{\bf Characteristics about the human:}\\
1. The human may send images and/or text to the assistant.\\
2. The human may ask questions requiring visual reasoning and/or understanding the relations between multiple images.\\
3. The human may ask the assistant to show images based on his/her intention.\\
4. The human may ask the assistant to do interesting things, rather than simply describing the content of the image.\\

{\bf Properties of a bad dialogue:}\\
1. Simply describing or analyzing the content in the image.\\
2. Dialogue without a good logic.\\

{\bf Properties of a good dialogue:}\\
1. Introducing extrinsic and attractive information of stuff shown in the image.\\
2. Discovering the connection between multiple images.\\
3. The dialogue happens in a reasonable context.\\

{\bf Examples:}\\
========\\
Case 1:\\
Input Images: \dots \\
Output Dialogue: \dots \\
---------\\
Case 2:\\
Input Images: \dots \\
Output Dialogue: \dots \\
---------\\
Case 3:\\
Input Images: \dots \\
Output Dialogue: \dots \\
======\\

Please directly give the dialogue if you understand. {\color{forestgreen}The number of turns of the dialogue should be less than 6.} The dialogue should be self-contained. {\color{blue} Do NOT} assume any previous dialogue between the human and the assistant. Please use the same format {\color{red}$<$imgX$>$ DESCRIPTION $<$/imgX$>$} to denote images in the dialogue and do not modify the description of the image. Note that the dialogue can stop at any time without an explicit ending.
\hfill
\end{AIbox}}
\caption{Illustration of the prompt used for \ours. \label{fig:app_gpt4_prompt}}
\end{figure*}
\section{Examples of Constructed Conversations \label{app:conv_examples}}
The examples are given in Fig. \ref{fig:app_conv_examples}.

\begin{figure*}
\centering
\resizebox{0.9\textwidth}{!}{
\begin{AIbox}{GPT-4 Ouput}
{\bf Example \#1}\\

{\bf Human}: Hi! I'm trying to write {\color{forestgreen} \bf a cartoon story} for my kids featuring a grumpy clown and a lovable cupcake. Can you help me craft this?\\
{\bf Assistant}: Of course! That sounds like a wonderful idea. We can start by establishing our characters. The grumpy clown could be someone who struggles finding joy in his own performances. {\color{red}$<$img0$>$ a cartoon illustration of a clown looking angry $<$/img0$>$} \\
{\bf Human}: That’s a wonderful idea! How about {\color{forestgreen} \bf the cute cupcake}?\\
{\bf Assistant}: The cupcake, on the other hand, is always upbeat and tries to help the clown find his happiness. Despite being a tiny cupcake, it has a big heart and a contagious joy. {\color{blue}$<$img1$>$ cartoon illustration of a cupcake with a happy expression $<$/img1$>$}\\
{\bf Human}: Interesting! What should be their adventure?\\
{\bf Assistant}: They could go on a journey to find the ultimate joke that guarantees laughter. The lovable cupcake, with his relentless positivity and cheer, helps the grumpy clown see the lighter side of life and finally find his joy.\\

{\bf Example \#2}\\

{\bf Human}: Hi, I am researching small business operations. Can you show me an example of informal small business?\\
{\bf Assistant}: Of course. Here's an example of informal small business: {\color{red}$<$img0$>$ a mother and daughter selling gum and cigarettes in person $<$/img0$>$}. This concept of business is most commonly found in rural or underdeveloped areas where formal establishments might not be as prevalent.\\
{\bf Human}: Interesting! On a somewhat related note, what sells well in such arrangements?\\
{\bf Assistant}: It depends on the location and the demand. Mostly, such businesses sell products of daily use such as food items, candy, toiletries etc. For example an ice cream cart is a common sight in many areas, as it's a popular treat among people of all ages.\\
{\bf Human}: Can you show me an image of a kid indulging in something like that?\\
{\bf Assistant}: Absolutely! Here's a delightful image of {\color{blue}$<$img1$>$ a child eating ice cream $<$/img1$>$}. Ice cream and similar snacks are often popular products for small informal businesses.\\

{\bf Example \#3}\\

{\bf Human}: Hi there! I'm preparing a speech on unity and cooperation in global issues. I have a few images but I'm not sure how to describe them in a way that fits my theme.\\
{\bf Assistant}: I'd be happy to help with that! Could you show me the images?\\
{\bf Human}: Sure, here they are. {\color{red}$<$img0$>$ large group of people in the shape of flag $<$/img0$>$} and {\color{blue}$<$img1$>$ rear view of a male boxer holding globe with flag painted on his back $<$/img1$>$}\\
{\bf Assistant}: The first image can be seen as a symbol of unity. Here, numerous individuals come together to form the shape of a flag, showing that collective action can lead to a powerful representation, like a flag - often a symbol of identity and unity. The second image illustrates the idea of global responsibility and commitment. The male boxer, suggesting strength and determination, has a globe – representing the world – in his hands, which could indicate a keen sense of global responsibility. The flag painted on his back can symbolize allegiance or commitment to his country, and by extension to global harmony if we're expanding the theme to encompass international cooperation.\\
{\bf Human}: That sounds great! I also have this image. Not sure if it fits. {\color{blue}$<$img2$>$ diplomatic handshake between countries : flags overprinted the hands stock photo $<$/img2$>$}\\
{\bf Assistant}: This image very much fits your theme. It can represent a formal agreement or cohesion between countries. The flags overprinted on the shaking hands signify the countries involved in the meet. This image can be used to illustrate diplomatic dialogue or mutual understanding which is an integral component to solving global issues through cooperation and unity.
% \hfill
\end{AIbox}}
\caption{Illustration of in-context examples of \ours. \label{fig:app_conv_examples}}

\end{figure*}
\section{Comparison with previous datasets.\label{app:cmp_data}}
We further investigate the lexical diversity of generated data, including both instructions and responses. The diversity of text in a dataset is defined as $\sum_{n=2}^{4}(\frac{\# \text{Unique n-grams}}{\# \text{Total n-grams}})$, in line with previous works \citep{su2022a}. As shown in Tab. \ref{tab:text_div}, our dataset achieves better diversity scores than most existing visual-language datasets, indicating that the language used in our dataset is more informative.

\begin{table*}
    \small
    \centering
    % \label{tab:subtable1}
    % \resizebox{0.9\textwidth}{!}{
    \begin{tabular}{lccccc}
    \toprule
    
    \multirow{2}{*}{\textbf{Dataset}} &  \multirow{2}{*}{\textbf{Multi-turn}} & \multirowcell{2}{\textbf{Arbitrarily} \\ \textbf{Interleaved}}  & \multirow{2}{*}{\textbf{Multimodal}} & \multirowcell{2}{\textbf{Instruction} \\ \textbf{Following}} & \multirowcell{2}{\textbf{Image} \\ \textbf{Generation}}\\
    &&&&&
    \\ \midrule
    \textsc{Platypus}       &\xmark     &\xmark     &\xmark      &\cmark &\xmark\\
    \textsc{Flamingo}       &\xmark     &\cmark &\cmark  &\xmark     &\cmark\\
    \textsc{MultiInstruct}  &\xmark     &\xmark     &\cmark  &\cmark &\xmark\\
    \textsc{MiniGPT-4}      &\xmark     &\xmark     &\cmark  &\cmark &\xmark \\
    \textsc{LLaVA}          &\cmark &\xmark     &\cmark  &\cmark &\xmark \\\midrule
   %\textsc{Shikra}         &\xmark     &\xmark     &\cmark  &\checkmark &\xmark\\
    \ours                   &\cmark &\cmark &\cmark  &\cmark &\cmark\\

    \bottomrule
    \end{tabular}
    % }
    \caption{Comparison with previous datasets.}
    \label{tab:cmp_data}
\end{table*}

\begin{figure*}[t]
    \begin{center}
        \includegraphics[width=1.0\linewidth]{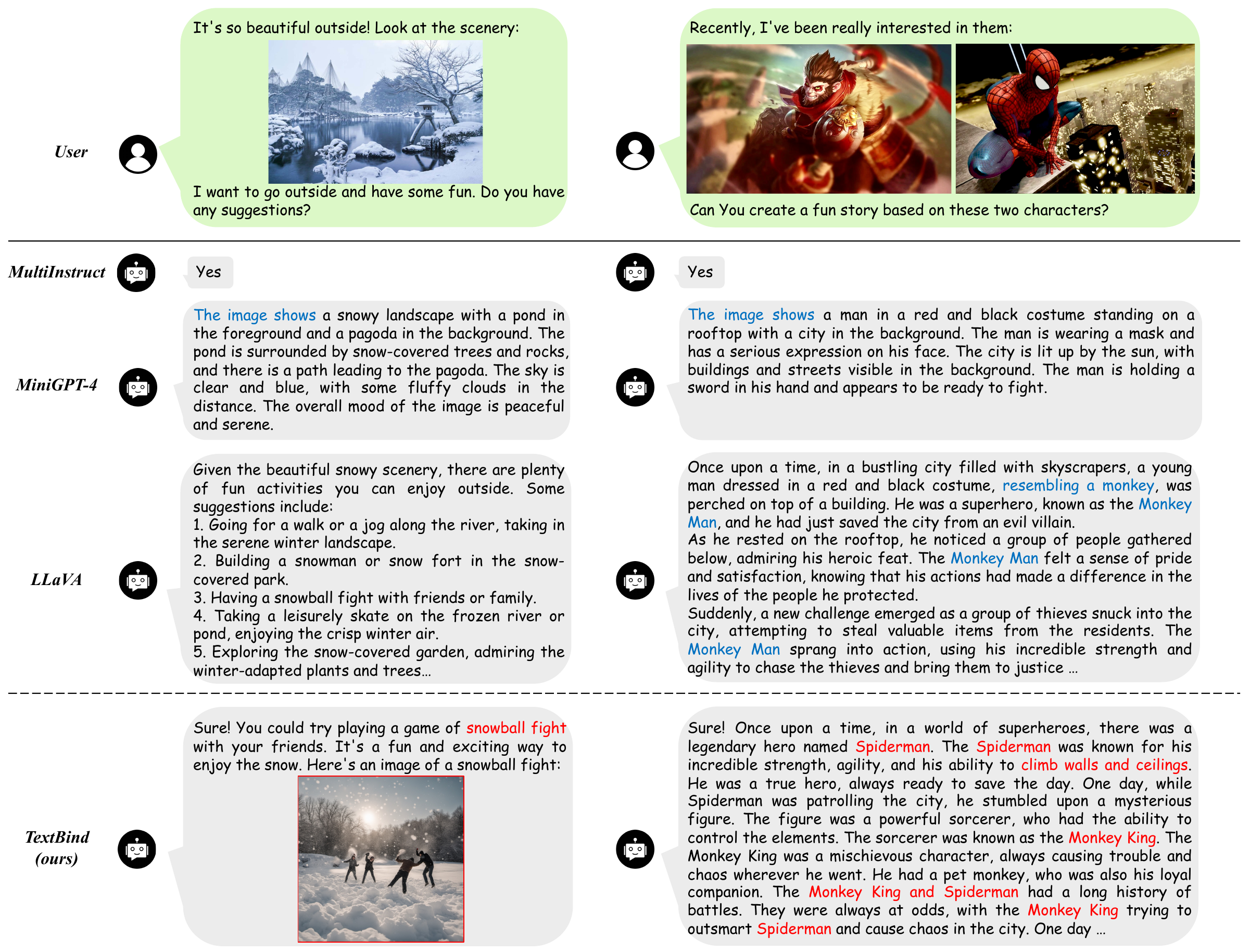}
    \end{center}
        \caption{Comparison among models trained on different datasets.
        \label{fig:cmp_model}}
\end{figure*}
\section{Human annotation guideline.\label{app:annotation}}
The comprehensive guideline for human evaluation is shown in Table \ref{tab:annotation}.

\begin{table*}[h!]
\centering
\resizebox{1.0\textwidth}{!}{
\begin{tabular}{c|l|p{0.7\textwidth}}
\toprule
\textbf{Annotation} & \textbf{Labels} & \textbf{Description} \\\midrule
\multirow{3}{*}{Overall Quality} & Excellent & This conversation is very interesting, practical, or intricate.  \\
 & Satisfactory & This conversation is coherent and reasonable without any factual errors.  \\
 & Poor  & At least one turn in the conversation is unreasonable in some aspects, e.g., unrealistic content, illegal formats, etc. \\\midrule
\multirow{4}{*}[-5ex]{Required Abilities} & Image Creation & To create new images in appropriate contexts. \\
 & Image Comparison & To combine, relate, or compare the information in different images.\\
 & Intrinsic Image Understanding  & To identify and recognize the objects, colors, shapes, and patterns in images.  \\
 & Extrinsic Image Understanding  & To interpret the underlying meaning of images, e.g., the context, emotions, symbolism, or narrative conveyed by the images. It goes beyond merely recognizing the elements in the images and often requires external knowledge and/or deep analysis.    \\
\bottomrule
\end{tabular}
}
\caption{Human annotation guideline.\label{tab:annotation}}
% \vspace{-2em}
\end{table*}

\section{Implementation Details (Data)\label{app:data_implementation_details}}
We construct our \ours dataset based on the \textsc{Conceptual Captions 3M} (CC3M) \citep{DBLP:conf/acl/SoricutDSG18, DBLP:conf/cvpr/ChangpinyoSDS21} dataset, which only provides image-caption pairs. In our experiments, we employ the \texttt{clip-vit-base-patch16} model released by OpenAI\footnote{\url{https://huggingface.co/openai/clip-vit-base-patch16}} \citep{pmlr-v139-radford21a} to filter out image-caption pairs with matching scores lower than $30$. We use the $k$-means clustering algorithm implemented by \texttt{FAISS} \citep{johnson2019billion} toolkit to classify the cleaned CC3M dataset into 4096 clusters. The features used for $k$-means clustering are the hidden representations of images encoded by \texttt{clip-vit-base-patch16} model. In addition, clusters with less than $32$ images are regarded as outliers and will not be considered. The number of images in each conversation is sampled from $\{2,3,4\}$. We access the GPT-4 model through the OpenAI API\footnote{\url{https://openai.com/blog/openai-api}}, and set \texttt{top\_p} and \texttt{temperature} hyper-parameters to $1.0$.
\section{Constructed Conversations with ``Poor'' Label\label{app:bad_example}}
In Table \ref{tab:error}, we identify three typical errors present in the constructed dataset. Despite setting a high threshold to filter out mismatched image-caption pairs, some mismatched cases cannot be detected by the CLIP model \citep{pmlr-v139-radford21a}. A few conversations suffering from incoherence and hallucinations may be attributed to the GPT-4 model. Overall, while a small number of conversations are affected by errors that are difficult to detect using rules, most generated conversations exhibit high quality. We present several cases labeled with ``Poor''. We can find that most of those ``Poor'' cases only have minor and non-obvious problems.

\begin{table}[h!]
\centering
\begin{tabular}{lc}
\toprule
\textbf{Error Type} & \textbf{Percentage}  \\\midrule
Img-Cap Mismatch  & $0.03$ \\
Incoherence & $0.03$ \\
Hallucination & $0.03$ \\
\bottomrule
\end{tabular}
\caption{Error types and percentage in dataset constructed by \ours.\label{tab:error}}
%\vspace{-1.0em}
\end{table}

\begin{figure*}
  \centering
  % First row
  \begin{subfigure}[b]{0.42\linewidth}
    \includegraphics[width=\textwidth]{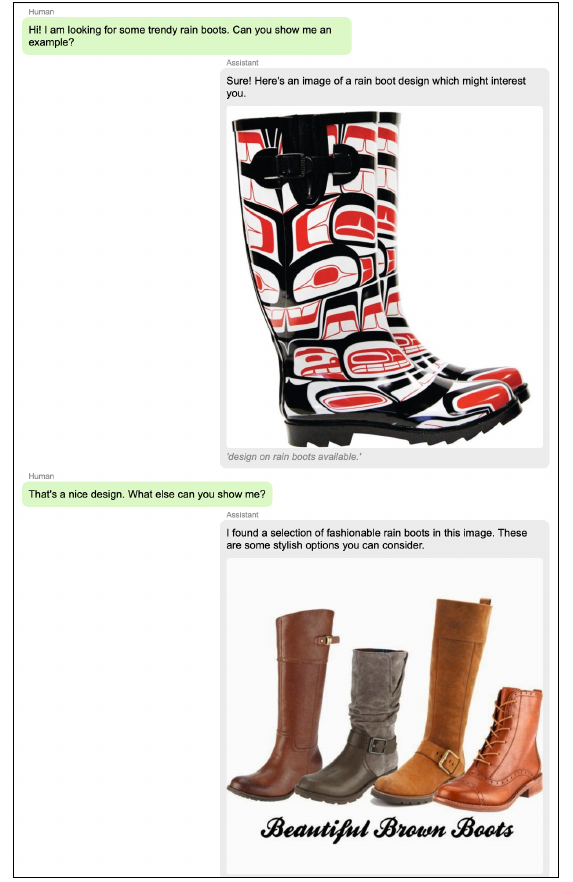}
    \caption{Hallucination Case: The topic is about rain boot. However, not all the boots shown in the second image are rain boots.}

  \end{subfigure}
  \hfill
  \begin{subfigure}[b]{0.42\linewidth}
    \includegraphics[width=\textwidth]{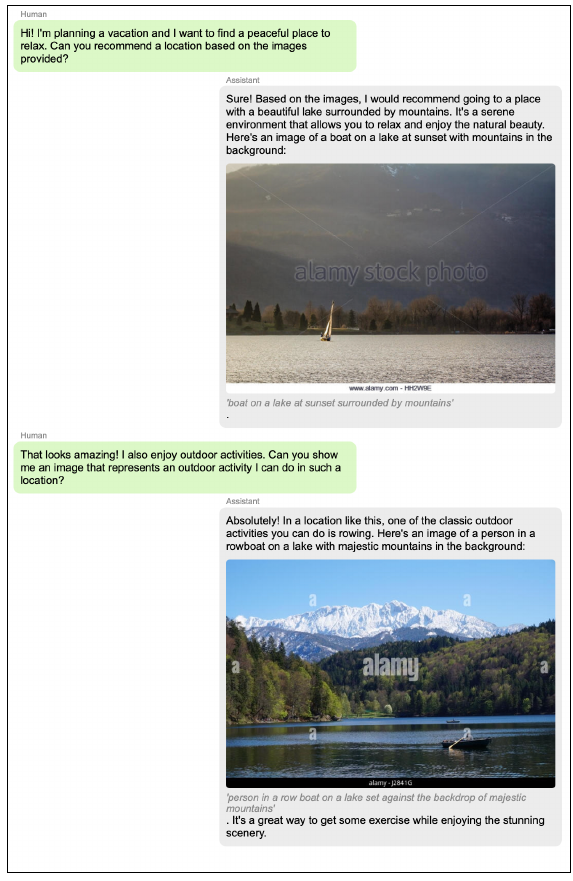}
    \caption{Incoherence Case: The first turn discusses about relaxing while the second turn is about outdoor activities. There are no connections.}

  \end{subfigure}

  \begin{subfigure}[b]{0.43\linewidth}
    \includegraphics[width=\textwidth]{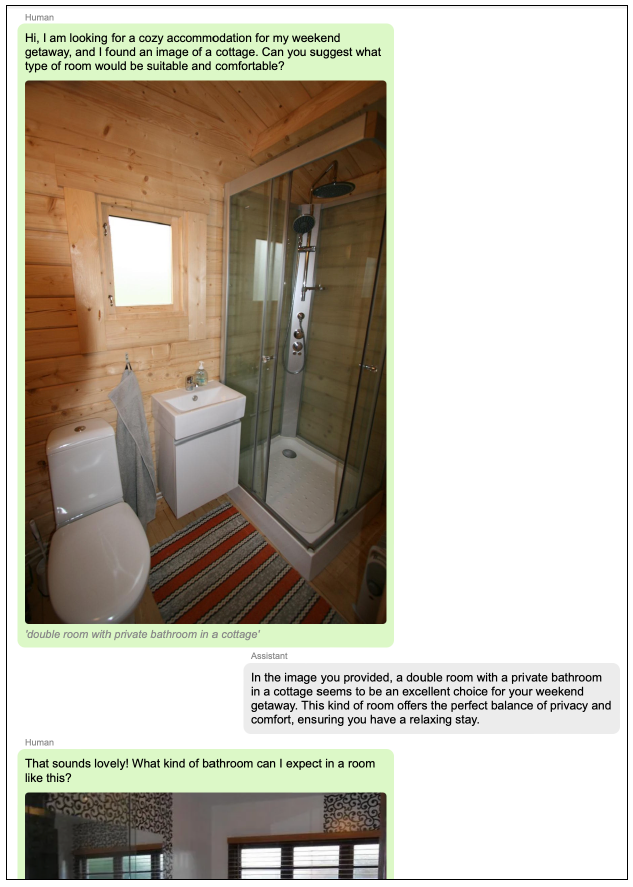}
    \caption{Image-caption Mismatch Case: The first image only shows a bathroom, but the caption is ``\textcolor{gray}{double room with private bathroom in a cottage}''. }

  \end{subfigure}

  \caption{Constructed conversations with ``Poor'' Label. The caption is shown below the image with \textcolor{gray}{gray color}.}
  \label{fig:poor}
\end{figure*}
\section{Implementation Details (Model)\label{app:model_implementation_details}}

\begin{table*}[h!]
  \centering
  \resizebox{0.95\linewidth}{!}{%
  \begin{tabular}{lccccc}
    \toprule
    \textbf{Training Stage} & \textbf{Epoch} & \textbf{Learning Rate} & \textbf{Batch Size} & \textbf{Max Sequence Length} & \textbf{Training Modules}  \\
    \midrule
    Multimodel Alignment & 2 & 1e-4 & 256 & 256 & Q-Former, Linear \\
    Multimodel Instruction Following & 3 & 1e-5 & 64 & 768 & Q-Former, Linear, LLM \\
    \bottomrule
  \end{tabular}
  }
  \caption{Training Configures of our Experiments}
  \label{tab:training_configures}
\end{table*}

Our experiments are based on Huggingface Transformers\footnote{\url{https://huggingface.co/docs/transformers/index}} \citep{DBLP:conf/emnlp/WolfDSCDMCRLFDS20} and DeepSpeed\footnote{\url{https://github.com/microsoft/DeepSpeed}} \citep{DBLP:conf/kdd/RasleyRRH20}.
We use the filtered synthetic captions given by BLIP\footnote{\url{https://github.com/salesforce/BLIP}} \citep{li2022blip}, including Conceptual Captions \citep{DBLP:conf/cvpr/ChangpinyoSDS21, DBLP:conf/acl/SoricutDSG18} and SBU \citep{DBLP:conf/nips/OrdonezKB11}, totally 12M image-caption pairs.
We employ the same visual encoder and Q-Former as used in BLIP-2 \citep{li2023blip} and use their weights for initialization.
LLama2-Chat\footnote{\url{https://huggingface.co/meta-llama/Llama-2-7b-chat-hf}} \citep{touvron2023llama} is utilized as the backbone language model.
For the image generation model, we use Stable Diffusion XL\footnote{\url{https://github.com/Stability-AI/generative-models}} \citep{podell2023sdxl}.
The training configures are shown in Table \ref{tab:training_configures}.
We use 8 NVIDIA A100 (40G) GPUs for all experiments.
\section{Human Evaluation Guideline \label{app:human_evaluation}}
For overall quality assessment, we adopt the following criteria.
\begin{itemize}
    \item Score 4: The response is excellent.
    \item Score 3: The response is acceptable but may not be very informative and interesting.
    \item Score 2: The response has minor issues, such as slight hallucinations when describing the images in the context.
    \item Score 1: The response is invalid and has significant drawbacks, e.g., irrelevant to the context.
\end{itemize}
For fine-grained annotations, intent recognition evaluates whether the response fulfills the the users' intent,  context understanding examines whether the response correctly comprehends the information in text and images, and the response informativeness measures the usefulness of the response. For each dimension, a human annotator will assign a score in $\{1,2,3,4\}$. The four scores ranging from 1 to 4 indicate ``major error'', ``minor error'', ``acceptable'', and ``perfect'', respectively.
\section{Data Format of Model Input \label{app:llama_data_struct}}
The data format of model input is given in Fig. \ref{fig:app_data_format}.
{
\color{red}{
\lstdefinelanguage{json}{
  basicstyle=\normalfont\ttfamily,
  numbers=left,
  numberstyle=\scriptsize,
  stepnumber=1,
  numbersep=8pt,
  showstringspaces=false,
  breaklines=true,
  frame=lines,
  backgroundcolor=\color{gray!10},
  stringstyle=\color{blue},
  morestring=[b]",
  morestring=[d]'
}
\lstset{
  basicstyle=\ttfamily\small, % Set the font and size of the code
  backgroundcolor=\color{gray!10}, % Set the background color of the code block
  keywordstyle=\color{blue}, % Set the color of keywords
  stringstyle=\color{red}, % Set the color of strings
  commentstyle=\color{green}, % Set the color of comments
  breaklines=true, % Break long lines
  showstringspaces=false, % Don't show spaces in strings
  numbers=left, % Show line numbers on the left
  numberstyle=\tiny\color{gray}, % Set the font and color of line numbers
}
\begin{figure*}
\begin{lstlisting}[language=json]
{
    "conversation" {
        {
            "role": "user",
            "content": "<image> Do you...",
            "image_list": ["1235.png"]
            "caption_list": ["a runing husky ..."]
        },
        {
            "role": "assistant",
            "content": "Yes, I do! ...",
            "image_list": [],
            "caption_list": []	
        },
        ...	
    }
}
\end{lstlisting}
\caption{Data Format of Model Input. \label{fig:app_data_format}}
\end{figure*}
}
}

\section{\ours Usage}
\ours is completely open-source and can be used for academic research and commercial purposes.

\end{document}